\documentclass[10pt,twocolumn,letterpaper]{article}
\usepackage{titlesec} 

\usepackage{iccv}              

\usepackage{subcaption}
\usepackage{graphicx}
\usepackage{caption}
 \usepackage{array}
\usepackage{helvet}
\usepackage{hhline}

\definecolor{iccvblue}{rgb}{0.21,0.49,0.74}
\usepackage[pagebackref,breaklinks,colorlinks,allcolors=iccvblue]{hyperref}

\def\paperID{14003} 
\def\confName{ICCV}
\def\confYear{2025}

\title{2HandedAfforder: Learning Precise Actionable Bimanual Affordances from Human Videos}

\author{Marvin Heidinger$^{*1}$, Snehal Jauhri$^{*1}$, Vignesh Prasad$^{1}$, Georgia Chalvatzaki$^{1,2}$\\
$^{*}$ indicates equal contribution\\
$^{1}$Computer Science Department, Technische Universit\"at Darmstadt, Germany \\ 
$^{2}$Hessian.AI, Darmstadt, Germany \\
\{\texttt{marvin.heidinger, snehal.jauhri, vignesh.prasad\}@tu-darmstadt.de},
\texttt{georgia.chalvatzaki@tu-darmstadt.de}
}

\author{Marvin Heidinger$^{*1}$, Snehal Jauhri$^{*1}$, Vignesh Prasad$^{1}$, Georgia Chalvatzaki$^{1,2}$\\
$^{*}$ indicates equal contribution\\
$^{1}$Computer Science Department, Technische Universit\"at Darmstadt, Germany\\
$^{2}$Hessian.AI, Darmstadt, Germany\\
{\tt\small \{snehal.jauhri, vignesh.prasad, georgia.chalvatzaki\}@tu-darmstadt.de}
}

\begin{document}
\maketitle

\begin{abstract}

When interacting with objects, humans effectively reason about which regions of objects are viable for an intended action, i.e., the affordance regions of the object. They can also account for subtle differences in object regions based on the task to be performed and whether one or two hands need to be used. However, current vision-based affordance prediction methods often reduce the problem to naive object part segmentation. In this work, we propose a framework for extracting affordance data from human activity video datasets. Our extracted 2HANDS dataset contains precise object affordance region segmentations and affordance class-labels as narrations of the activity performed. The data also accounts for bimanual actions, i.e., two hands co-ordinating and interacting with one or more objects. We present a VLM-based affordance prediction model, 2HandedAfforder, trained on the dataset and demonstrate superior performance over baselines in affordance region segmentation for various activities. Finally, we show that our predicted affordance regions are actionable, i.e., can be used by an agent performing a task, through demonstration in robotic manipulation scenarios. Project-website:~\href{https://sites.google.com/view/2handedafforder}{sites.google.com/view/2handedafforder}
\end{abstract}

\vspace{-15pt}

\section{Introduction}
\label{sec:intro}

\begin{figure}
    \centering
    \includegraphics[width=1.0\columnwidth]{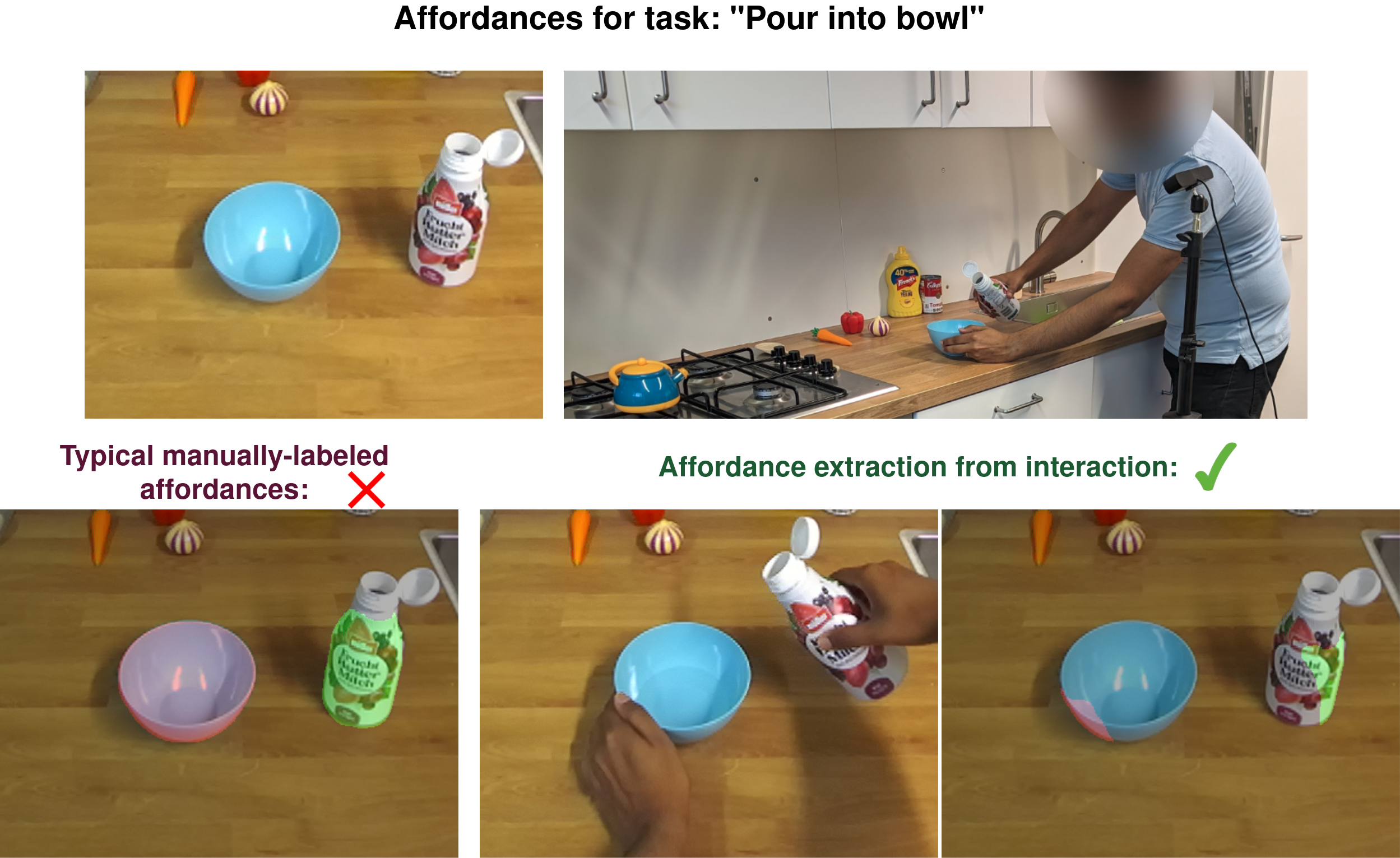}
    \captionof{figure}{\textbf{A motivating example:} When labeling affordances for a task `Pour into bowl', typical labeled affordances provided by annotators are not precise and reduce the problem to object part segmentation. Alternatively, our affordance extraction method uses the hand-object interaction sequence to get precise bimanual affordance regions that are not just ‘conceptual’ but also ‘actionable’.
    }
    \label{fig:cover}
    \vspace{-1.5em}
\end{figure}
When humans perceive objects, they understand different object regions and can predict which object region \textit{affords} which activities~\cite{gibson2014theory}, i.e., which object regions can be used for a task. We wish our machines to have this ability, referred to in literature as ``affordance grounding''. Affordance grounding has several downstream applications, including building planning agents, VR, and robotics. 
Affordance grounding is especially important for robotics since robots must reason about various actions that can be performed using different object regions which is a crucial step towards performing useful tasks in everyday, unstructured environments. For example, to pour into a bowl, the robot should know that it should hold the bottle in a region close to the center of mass of the bottle~(Figure~\ref{fig:cover}), i.e., a region that \textit{affords} pouring. Predicting such affordance regions is challenging since it requires a fine-grained understanding of object regions and their semantic relationship to the task.


\begin{table*}[ht!]
    \centering
    \resizebox{\textwidth}{!}{
    \begin{tabular}{|c|c|c|c|c|c|c|c|c|}
    \hline
         \textbf{Dataset} & \textbf{Image type \& source} & \textbf{\# Images} & \multicolumn{6}{c|}{\textbf{Affordance data}}\\
         & & &
         \textbf{Annotation source} & \textbf{Annotation type} & \textbf{\# Aff. classes} & \textbf{\# Obj. classes} & \textbf{Class-labels} & \textbf{Bimanual} \\
         \hline
         IIT-AFF~\cite{nguyen2017object} & Exocentric~\cite{russakovsky2015imagenet} & 8.8K & Manually-labeled & Masks & 9 & 10 & Explicit & No\\
         AGD20K~\cite{luo2022learning} & Exo+Egocentric~\cite{lin2014microsoft,chao2018learning} & 23.8K & Manually-labeled & Heatmaps & 36 & 50 & Explicit & No \\         3DOI~\cite{qian2023understanding} & Exo+Egocentric~\cite{qian2022understanding, Damen2022RESCALING} & 10K & Manually-labeled & Points & 3 & n.a. & Explicit & No \\
         ACP~\cite{goyal2022human} & Egocentric~\cite{Damen2022RESCALING} & 15K & Auto-labeled & Heatmaps & n.a. & n.a. & none & No \\
         VRB~\cite{bahl2023affordances} & Egocentric~\cite{Damen2022RESCALING} & 54K & Auto-labeled & Heatmaps & n.a. & n.a. & none & No\\
         \textbf{2HANDS} & Egocentric~\cite{Damen2022RESCALING} & \textbf{278K} & Auto-labeled & \textbf{Precise Masks} & \textbf{73} & \textbf{163} & \textbf{Narrations} & \textbf{Yes} \\
         \hline
    \end{tabular}
    }
    \caption{Comparison of our dataset 2HANDS against other affordance prediction datasets. For 2HANDS, we auto-label a large number of affordance region masks from human egocentric videos and use narration-based affordance class-labels. Our dataset also contains bimanual masks, with the goal of addressing the challenging problem of precise bimanual affordance prediction.}
    \label{tab:my_label}
    \vspace{-1.0 em}
\end{table*}

Recent advances in large-language and multimodal models have shown impressive visual reasoning capabilities using self-supervised objectives~\cite{radford2021learning, openai2024gpt4technicalreport, deitke2024molmopixmoopenweights}. However, there is still a big gap in their ability to detect accurate object affordance regions in images~\cite{li2024one}. Moreover, most existing state-of-the-art affordance detection methods~\cite{guo2023handal, qian2023understanding, qian2024affordancellm, sun2023going, lai2024lisa} use labeled data~\cite{nguyen2017object, qian2023understanding, luo2022learning, he2022partimagenet, lee2024cookar} that lacks precision and is more akin to object part segmentation rather than \textit{actionable} affordance-region prediction. When humans interact with objects, they are much more \textit{precise} and use specific object regions important in the context of the task. An example is provided in Fig.~\ref{fig:cover}. For the task of pouring into the bowl, part segmentation labels the entire bottom of the bottle with the affordance `pour'. But, to pour correctly, humans leverage the appropriate region of the bottle. Moreover, the affordances are inherently bimanual, i.e., the affordance regions of the bowl and bottle are interconnected.

We argue that affordances should not be labeled but automatically extracted by observing humans performing tasks, e.g. in activity video datasets. We propose a method that uses hand-inpainting and mask completion to extract affordance regions occluded by human hands. This has several advantages. First, by using this procedure, we are able to obtain \textbf{bimanual} and \textbf{precise} affordances~(Figure ~\ref{fig:cover}) rather than simply predicting object parts. Second, 
it makes affordance specification more natural since it is often easier for humans to \textit{show} the object region to interact with, rather than label and segment it correctly in an image. Third, 
using human activity videos gives us diverse task-specific affordances, with the affordance class label naturally coming from the narration of what task is being done by the human. This makes our affordances \textbf{task-oriented} with natural language specification, unlike previous methods focused on predicting task-agnostic interaction hotspots~\cite{goyal2022human, bahl2023affordances}.

We extract a dataset, 2HANDS~(2-Handed Affordance + Narration DataSet), consisting of a large number of unimanual and bimanual object affordance segmentation masks and task narrations as affordance class-labels. We propose a VLM-based affordance prediction model, 2HandedAfforder, that is trained on the 2HANDS dataset and predicts affordance masks in images based on an input text prompt. To evaluate the performance on this challenging problem, we also present a novel benchmark, ActAffordance, using annotations on images from two egocentric human activity datasets~\cite{Damen2022RESCALING, grauman2022ego4d}. Our contributions are:
\begin{itemize}
	\item a method to extract precise affordance regions from human-object interaction videos.
	\item a dataset, 2HANDS, consisting of 278K images with extracted affordance masks, narration-based class labels, and unimanual/bimanual taxonomy labels.
    \item an affordance network, 2HandedAfforder, for predicting 
    task-aware unimanual and bimanual affordance regions.
    \item a new benchmark, ActAffordance, with affordance annotations by humans who observe the interaction sequence.
    \item the first comprehensive dataset and evaluation of task-specific bimanual object affordance regions in images.
\end{itemize}


\section{Related Work}

\textbf{Fully supervised affordance detection.} In fully supervised affordance detection datasets and methods such as by \citet{qian2023understanding}, AffordanceLLM~\cite{qian2024affordancellm}, the dataset is fixed and hand-annotated such as from IIT-AFF~\cite{nguyen2017object} and 3DOI~\cite{qian2023understanding}. The affordance classes in these datasets are explicit and annotators guess which affordance class may apply to object regions. Other methods, such as VLPart~\cite{sun2023going}, use a general open vocabulary segmentation pipeline. LISA~\cite{lai2024lisa} performs open-vocabulary, prompt-based ``reasoning segmentation". However, these methods do not consider actions and typically segment either the whole object~\cite{lai2024lisa} or object parts\cite{sun2023going}, and not precise affordance regions.

\textbf{Weakly supervised affordance detection.} Weakly supervised methods such as Cross-viewAG~\cite{luo2022learning} and Locate~\cite{li2023locate} learn to predict affordances by observing exocentric images of humans interacting with objects based on the AGD20K dataset~\cite{luo2022learning}. The model maps object parts across images, transferring the learned affordances to non-exocentric images where no hand-object interaction occurs. This is similar to saliency matching methods that use one-shot affordance transfer~\cite{zhai2022one, hadjivelichkov2023one}. However, these methods still require an initial smaller manually-labeled dataset with explicit affordance classes.

\begin{figure*}[t!]
    \centering
    \includegraphics[width=\textwidth]{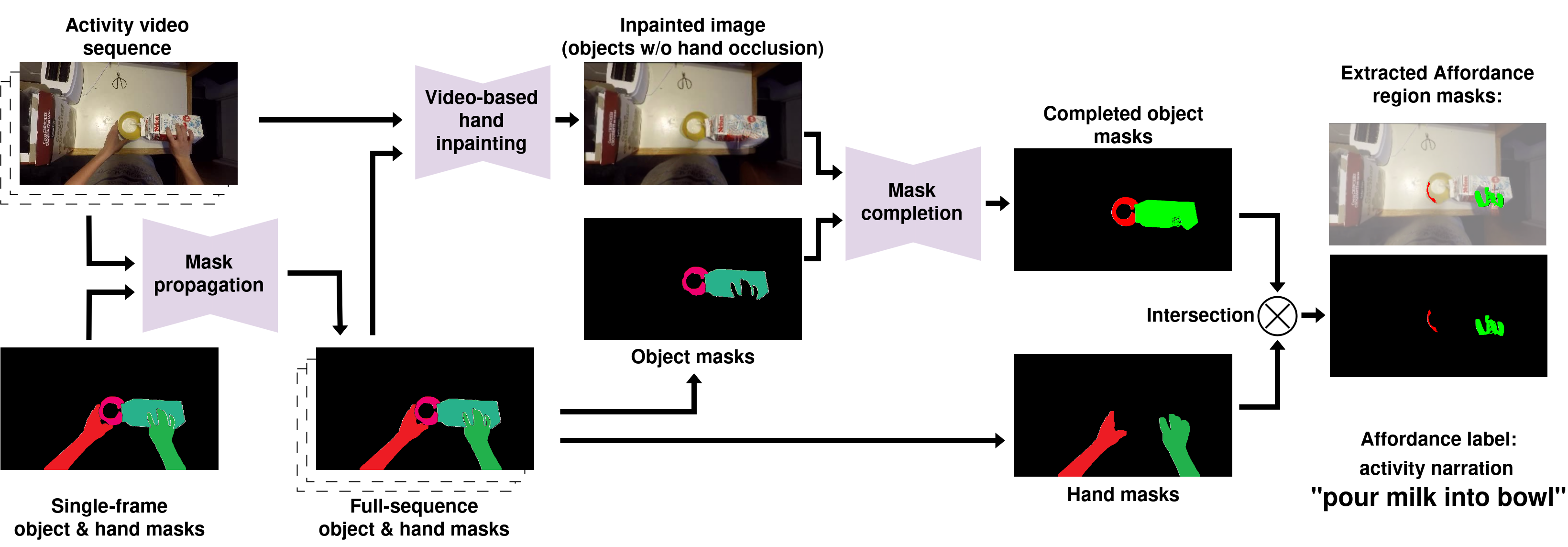}
    \caption{Affordance extraction pipeline. Given a human activity video sequence and a single-frame object and hand masks, we first obtain dense, full-sequence object and hand masks using a video mask-propagation network~\cite{cheng2022xmem}. We then inpaint out the hands in the RGB images using a video-based hand inpainting model~\cite{chang2024look}. This gives us an image with the objects reconstructed and un-occluded by the hands. With the inpainted image and the original object masks, we use~\cite{ravi2024sam} to ``complete" the object masks by again propagating the object masks to the inpainted image. Finally, we can extract the affordance region masks for the given task as the intersection between the completed masks and the hand masks. We also label the affordance class using the narration of the task.\vspace{-1.5em}}
    \label{fig:aff_extract}
\end{figure*}

\textbf{Auto-labeled affordance detection.}
Egocentric videos of humans performing tasks~\cite{Damen2022RESCALING, VISOR2022, grauman2022ego4d, zhang2022fine, grauman2024ego} are an attractive option for extracting affordance data since they include object interactions up close and in the camera field of view. Recently,~\citet{goyal2022human} and \citet{bahl2023affordances} have shown that videos from datasets such as EPIC kitchens~\cite{Damen2022RESCALING} and Ego4D~\cite{grauman2022ego4d} can be used to segment regions of interest in objects using weak supervision from hand and object bounding-boxes. However, these works focus on segmenting task-agnostic `hotspot' interaction regions of objects. The region of interactions do not consider the task and whether one or two hands would be needed.

\textbf{Our approach and goals.}
In this work, we propose a method to extract affordance masks leveraging recent video-based hand inpainting techniques~\cite{chang2024look}. 
Since our dataset contains precise segmentation masks, we can predict pixel-wise affordance segments in the image, as opposed to methods only trained with point-labels of affordance~\cite{qian2023understanding} or that only predict heatmaps~\cite{luo2022learning, bahl2023affordances, srirama2024hrpSmall}. Moreover, we consider the especially challenging problem of bimanual affordance detection, for which the spatial context of the objects and their interconnection is also important. Although bimanual affordances have been considered in previous work~\cite{krebs2022bimanual, gorjup2019intuitive, sanchez2020motor, liu2024tacoSmall, plonka2024learning, fu2025gigahandsSmall}, to the best of our knowledge, ours is the first method to extract bimanual affordances from videos which we then use to train our model to predict task-specific affordance masks based on a text prompt.



\section{Extraction and Learning of Bimanual Affordances from Human Videos}
\label{sec:method}

In this section, we detail our affordance extraction approach used to generate our 2HANDS dataset from videos of humans performing everyday tasks (Sec.~\ref{ssec:affordance-extraction}). Then, we present our approach, ``2HandedAfforder'', for predicting meaningful task-oriented bimanual affordance regions in images in Sec.~\ref{ssec:affordance-prediction}.

\begin{figure*}[ht!]
    \centering
    \includegraphics[width=0.9\textwidth]{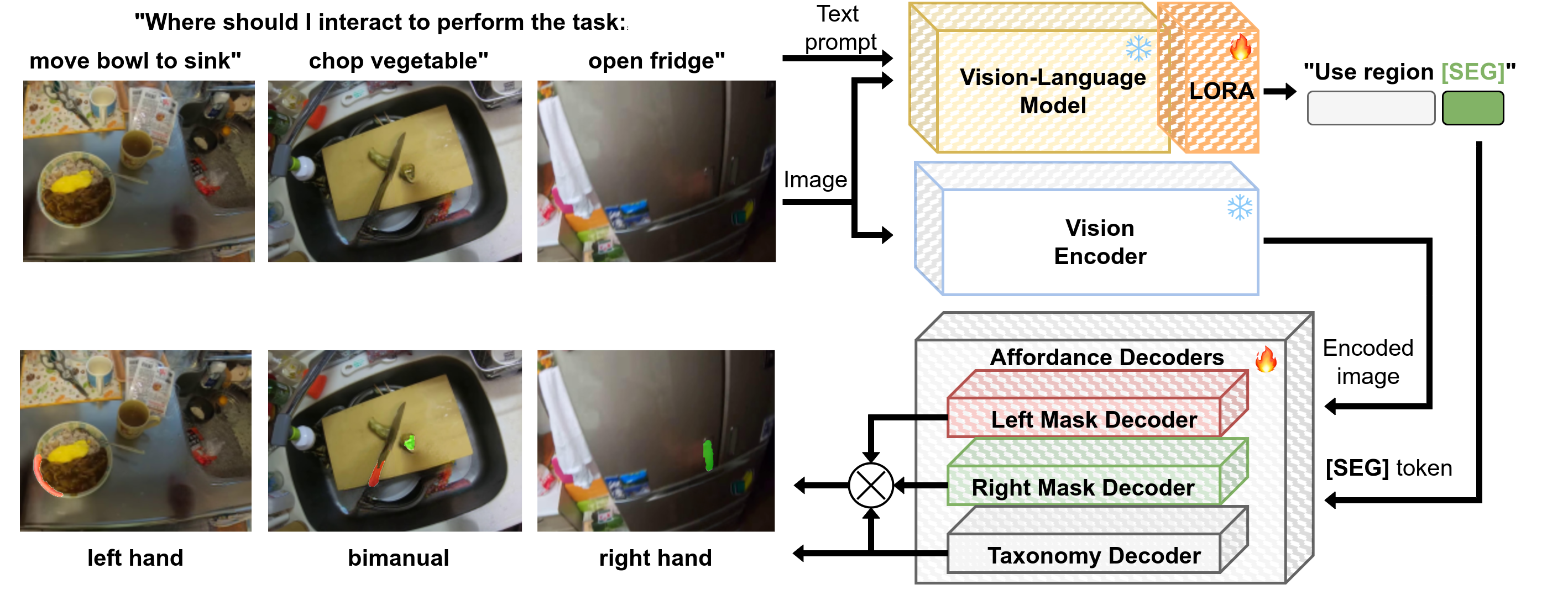}
    \caption{Affordance prediction network. Given an input image and task, we use a question asking where the objects should be interacted for the desired task as a text prompt to a Vision-Language model (VLM). The VLM produces language tokens and a [SEG] token which is passed to the affordance decoders. We also use a SAM~\cite{kirillov2023segany} vision-backbone to encode the image and pass it to the affordance decoders. The decoders predict the left hand and right hand affordance region masks as well as a taxonomy classification indicating whether the interaction is supposed to be performed with the left hand, right hand, or both hands. The vision encoder is frozen, while the VLM predictions are fine-tuned using LORA~\cite{hu2021lora}. \vspace{-1.5em}}
    \label{fig:model}
\end{figure*}

\subsection{Affordance Extraction from Human Videos}
\label{ssec:affordance-extraction}

We use videos of humans performing tasks to extract precise affordance masks. This involves closely examining the contact regions between the hands and objects in the videos. Several recent methods~\cite{shan2020understanding, potamias2024wilor} have shown impressive performance in hand-object segmentation and reconstruction. However, the challenge in affordance region extraction is that the hand typically occludes the object region with which it interacts. \citet{bahl2023affordances} circumvent this issue by only considering videos where objects are initially un-occluded before the interaction and only use the hand bounding-box to denote the interaction region. However, not only is this a limiting assumption, but also the bounding-boxes can only be used to detect interaction hotspots and not precise object affordance masks. Precise masks are more explicit and useful for downstream application, for example, for 
providing graspable regions of an object for robotic manipulation tasks. We propose a pipeline to extract affordances that leverages recent advances in hand inpainting~\cite{chang2024look} and object mask completion~\cite{sargsyan2023mi,ravi2024sam}, providing the first bimanual affordance region segmentation dataset. Moreover, we use the narration of the task being performed as the affordance text label, which helps obtain a diverse set of affordance classes for various objects. The full extraction pipeline is visualized in Figure~\ref{fig:aff_extract}.

We extract affordances from EPIC-KITCHENS~\cite{Damen2022RESCALING}, which contains $\sim$100 hours of egocentric videos of human activities in kitchens. We use the VISOR~\cite{VISOR2022} annotations of the dataset, which contain some sparse hand-object mask segmentations and binary labels denoting whether the hand is in contact with the object. Note that we can also use other video datasets like Ego4D~\cite{grauman2022ego4d} along with hand segmentation methods~\cite{shan2020understanding} to extract hand-object masks and contact/no-contact labels. 
To obtain dense hand-object masks for entire video sequences, we use a video-based mask propagation network~\cite{cheng2022xmem}.

With the hand and object masks available over the entire video sequence, we obtain an un-occluded view of the objects by inpainting out the hands. We use a video-based hand inpainting model, VIDM~\cite{chang2024look}, that uses 4 frames from the sequence as input to inpaint the missing regions. This sequence-based inpainting better reconstructs the target objects since the objects may be visible in another frame of the sequence without occlusion. Inpainting provides us with an un-occluded view of the objects. We then precisely segment these un-occluded objects in the inpainted image using mask completion. For this, we use the segmentation masks from the original image and prompt SAM2~\cite{ravi2024sam} to propagate these masks to the new inpainted image. We observe that this process gives us more precise object masks compared to directly using mask completion methods~\cite{sargsyan2023mi}, detailed in the appendix (Sec.~\ref{sec:mask-completion-approaches}).

To obtain the final affordance region where the hand interacted with the object, we can simply compute the intersection of the un-occluded object masks and the hand masks. The full pipeline is shown in Fig.~\ref{fig:aff_extract}. For bimanual affordances, it is also useful to classify the affordances into a bimanual taxonomy~\cite{krebs2022bimanual}. Thus, we distinguish between unimanual left, unimanual right, and bimanual actions. 
Additional details about the extraction procedure are provided in the appendix.

With the above procedure, we obtain a dataset of 278K images with extracted affordance segmentation masks, narration-based class-labels, and bimanual taxonomy annotations. We call this dataset 2HANDS, i.e., the \textbf{2}-\textbf{H}anded \textbf{A}ffordance + \textbf{N}arration \textbf{D}ata\textbf{S}et.


\subsection{Task-oriented Bimanual Affordance Prediction}
\label{ssec:affordance-prediction}
Reasoning segmentation, i.e., text-prompt-based segmentation of full objects, is a difficult task. Segmentation of precise object affordance regions is even more challenging. The complexity is further increased when considering bimanual affordances with multiple objects. To address this challenge, we develop a model for general-purpose bimanual affordance prediction that can process both an input image and any task prompt (e.g., "pour tea from kettle"). We call this model ``2HandedAfforder." 
We leverage recent developments in reasoning-based segmentation methods~\cite{liang2023open,lai2024lisa} and train a VLM-based segmentation model to reason about the required task and predict the relevant affordance region in the input image. Since our 2HANDS dataset provides precise segmentation masks, we can predict pixel-wise affordance segments in the image, as opposed to other methods that are only trained with point labels of affordance~\cite{qian2023understanding} or that only predict heatmaps~\cite{luo2022learning, bahl2023affordances}. 


Inspired by reasoning segmentation methods such as by \citet{lai2024lisa}, we use a Vision-Language Model~(VLM)~\cite{liu2024visual} to jointly process the input text prompt and image and produce language tokens and a segmentation [SEG] token as output. While VLMs excel at tasks such as visual question answering and image captioning, they are not explicitly optimized for vision tasks like segmentation, where accurately predicting pixel-level information is key. Therefore, to have a stronger vision-backbone for our segmentation-related task, we use a modified version of SAM~\cite{kirillov2023segany}. Given the combined embedding provided by the VLM [SEG] token and SAM image encoder, we use affordance decoders modeled after SAM-style mask decoders to predict the affordances. We use two mask decoders, generating two separate affordance masks for the left and right hands, respectively. Furthermore, we add a prediction head to one of the decoders that takes the output token as input and predicts the bimanual taxonomy: `unimanual left hand', `unimanual right hand', and `bimanual' using a separate full-connected classifier decoder. An overview of the whole network architecture is visualized in Figure~\ref{fig:model}.

The VLM is trained to generate a specific output token: a segmentation [SEG] token. Specifically, inspired by LISA~\cite{lai2024lisa}, we use question-answer templates to encapsulate the narration of the individual tasks in natural language, e.g. ``\textbf{USER:} $[$IMAGE$]$ Where would you interact with the objects to perform the action \{\textit{action\_narration}\} in this image? \textbf{ANSWER:} Use region: $[$SEG$]$.'' This [SEG] token encapsulates the general-purpose reasoning information from the VLM for the task which is then used by the affordance decoders. For the left and right hand mask decoders, we initialize the decoders with pre-trained SAM weights and train them to predict segmentation masks using the encoded image and [SEG] token as input. For the taxonomy classifier decoder, as in~\cite{qian2023understanding}, we pass the left mask decoder output token through an MLP to predict whether the action should be performed with the left hand, right hand, or both hands.


We freeze the weights of the image encoder and the VLM, and we apply Low-Rank Adaptation~(LoRA)~\cite{hu2021lora} to fine-tune the VLM. By introducing trainable low-rank updates, LoRA enables efficient fine-tuning of the VLM without requiring modifications to its original parameters. This ensures that the pre-trained knowledge of the VLM, a LLaVa-13b, is preserved while still allowing the model to specialize in segmentation. We do not fine-tune the SAM image encoder as this was shown to reduce performance in reasoning segmentation tasks. 
For training the mask prediction, we use a combination of dice loss~\cite{milletari2016v} and focal cross-entropy loss~\cite{ross2017focal}. For the taxonomy prediction, we use a cross-entropy loss with the ground truth label.
If the task does not require one of the hands, the weight for the corresponding mask loss is set to $0$. Similarly, when predicting affordance regions using the network at test time, we use the taxonomy prediction to infer whether left, right, or both mask predictions should be considered.

As an alternative to the VLM-based 2HandedAfforder prediction network, we also train a smaller CLIP-based~\cite{lin2023clip} version of the network that uses CLIP text features instead of the VLM [SEG] token as input to the affordance decoders. We call this network `2HandedAfforder-CLIP'.



\section{Experimental Setup}
\label{sec:exps}

With our experiments, we aim to answer the following questions:
\begin{enumerate}
    \item Does our affordance extraction procedure for the 2HANDS dataset provide accurate affordance region segmentation data?
    \item Is our 2HandedAfforder model able to predict precise unimanual and bimanual affordances? And how does it compare against baselines?
    \item How well does our affordance prediction model generalize to images in-the-wild?
    \item Are our affordances actionable, i.e., can they be utilized in real-world scenarios such as for robotic manipulation?
\end{enumerate}

\subsection{ActAffordance Benchmark}
\label{ssec:actaffordnace-benchmark}

To answer the first question of the accuracy of our extracted affordances in the 2HANDS dataset, we evaluate the alignment of our extracted affordance masks with human-annotated affordance regions. As mentioned in Sec.~\ref{ssec:affordance-extraction}, when humans label affordances, they often simply label object parts and do not necessarily focus on the precise regions of interaction of the objects~\cite{luo2022learning, qian2023understanding}. 
Moreover, the second question regarding the accuracy of 2HandedAfforder is non-trivial. Using only the masks in our 2HANDS dataset as ``ground truth" leads to a bias towards our own extracted affordances. Therefore, we propose a novel benchmark called ``ActAffordance'' to evaluate both the dataset quality and the predicted affordances. Specifically, we evaluate the alignment of our affordances with the affordances annotated by humans who are shown the interaction video sequence.

\begin{figure}[t!]
    \centering
    \begin{subfigure}[b]{0.4925\columnwidth}
        \centering
        \includegraphics[width=\columnwidth, trim={5cm 0 0 0},clip]{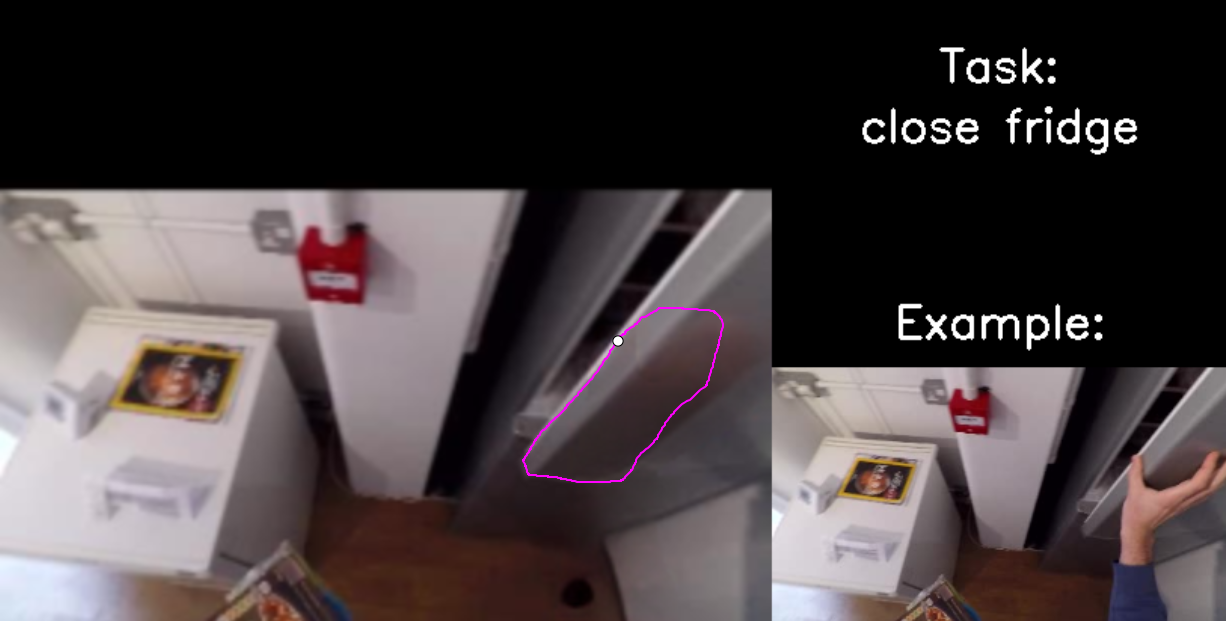}
    \end{subfigure}
    \hfill
    \begin{subfigure}[b]{0.4925\columnwidth}
        \centering
        \includegraphics[width=\columnwidth, trim={5cm 0 0 0},clip]{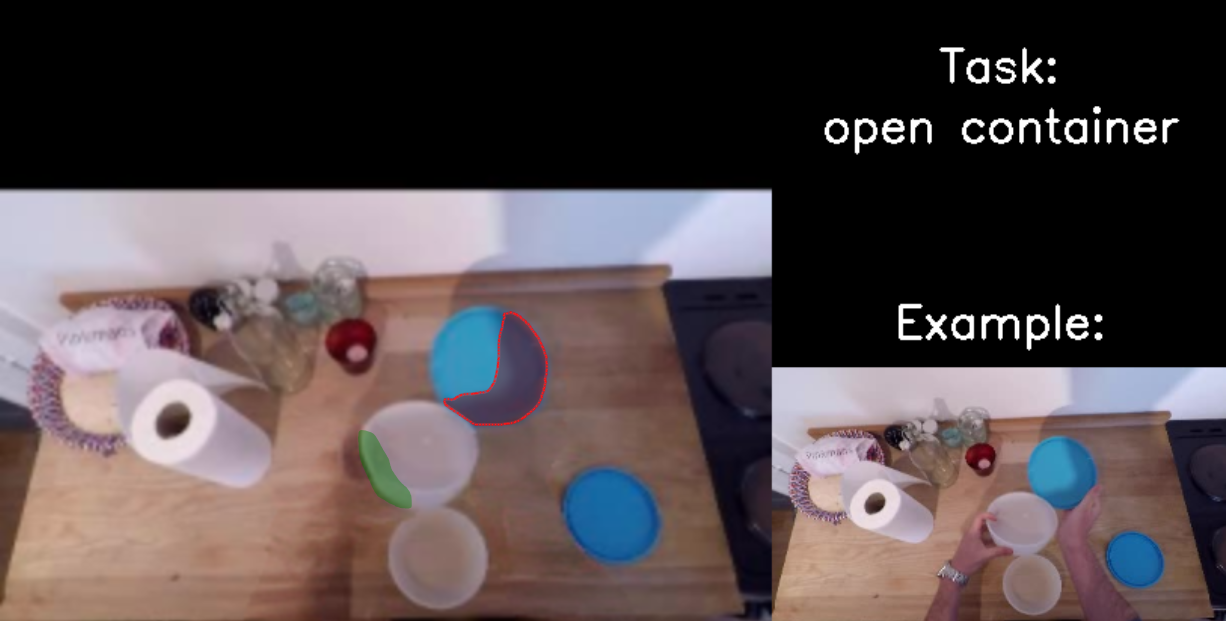}
    \end{subfigure}
    \caption{Example annotations for the ActAffordance benchmark. Left: The image to be annotated with the highlighted annotation mask(s). Right: the example interaction provided to the human annotator, along with the task description. The human is asked to annotate ALL the possible regions for the interaction to capture all the different modes. \vspace{-1.5em}}
    \label{fig:benchmark-annotation}
\end{figure}


For the ``ActAffordance'' benchmark, we asked 10 human annotators to label affordance regions with a novel approach: instead of direct segment labeling, we showed them pairs of inpainted and original hand-object interaction images. By \textit{showing annotators example interactions}, we asked them to predict similar affordance regions. Fig.~\ref{fig:benchmark-annotation} illustrates this annotation pipeline. Annotators predicted ALL possible interaction regions since affordance prediction is inherently multi-modal—for instance, when closing a fridge, a human might choose any point along the door length.
The benchmark contains unimanual and bimanual segmentation masks for 400 activities from EPIC-KITCHENS~\cite{Damen2022RESCALING} and Ego4D~\cite{grauman2022ego4d}, with no overlap between EPIC-KITCHENS data used in 2HANDS. Details about the benchmark and annotation process are in Appendix Sec.~\ref{sec:actaffordance-annotation}.

Another point of consideration when evaluating the affordance prediction is that the problem can be divided into two parts: correct identification of the objects based on the text prompt and accurate affordance region segmentation. Since these are two complementary but different capabilities, we further create another version of the benchmark called ``ActAffordance-Cropped". Here, we crop the benchmark images to a bounding box containing the target objects. This helps differentiate between the capabilities of segmenting the correct object and segmenting the correct object region. Moreover, it helps evaluate our network predictions against baselines that cannot identify correct objects in images but use bounding-boxes~\cite{bahl2023affordances} or query points on the object~\cite{qian2023understanding} as input.

We note that ActAffordance is a very challenging benchmark. To date, reasoning segmentation, i.e., text-prompt-based segmentation of full objects, is an unsolved problem. Prompt-based segmentation of precise object affordance regions is yet more challenging, especially when benchmarked against humans. The inclusion of bimanual affordances with multiple objects is another step beyond that. However, we feel this challenging benchmark will push the community forward towards more effective affordance prediction and thus we evaluate all methods on this benchmark instead of directly using the test set from our dataset.

\begin{figure}[t!]
    \centering
    \renewcommand{\arraystretch}{1.0} 
    \setlength{\tabcolsep}{1pt} 
    
    \begin{tabular}{@{}m{0.15\columnwidth}@{\hskip 1pt}m{0.28\columnwidth}@{\hskip 1pt}m{0.28\columnwidth}@{\hskip 1pt}m{0.28\columnwidth}@{}}
    
        &  \centering \renewcommand{\familydefault}{\sfdefault}     \fontfamily{phv}\selectfont\fontsize{6.5pt}{8pt}\selectfont\textbf{put lid on container} &  \centering \renewcommand{\familydefault}{\sfdefault}     \fontfamily{phv}\selectfont\fontsize{6.5pt}{8pt}\selectfont\textbf{put some water in the frying pan} &  \centering \renewcommand{\familydefault}{\sfdefault}     \fontfamily{phv}\selectfont\fontsize{6.5pt}{8pt}\selectfont\textbf{put down knife} \tabularnewline

         \centering \renewcommand{\familydefault}{\sfdefault}     \fontfamily{phv}\selectfont\fontsize{6.5pt}{8pt}\selectfont\textbf{LOCATE} & 
        \includegraphics[width=\linewidth]{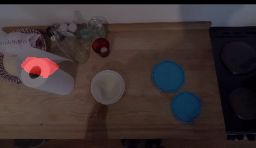} & 
        \includegraphics[width=\linewidth]{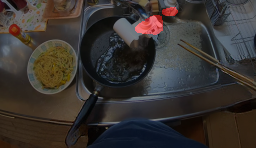} & 
        \includegraphics[width=\linewidth]{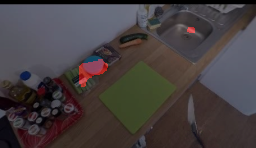} \tabularnewline
        
         \centering \renewcommand{\familydefault}{\sfdefault}     \fontfamily{phv}\selectfont\fontsize{6.5pt}{8pt}\selectfont\textbf{VRB} & 
        \includegraphics[width=\linewidth]{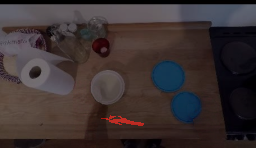} & 
        \includegraphics[width=\linewidth]{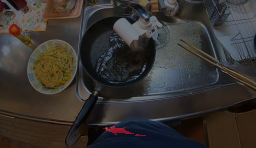} & 
        \includegraphics[width=\linewidth]{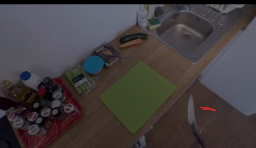} \tabularnewline
        
         \centering \renewcommand{\familydefault}{\sfdefault}     \fontfamily{phv}\selectfont\fontsize{6.5pt}{8pt}\selectfont\textbf{LISA} & 
        \includegraphics[width=\linewidth]{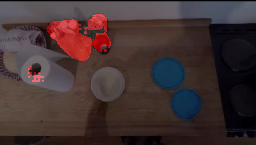} & 
        \includegraphics[width=\linewidth]{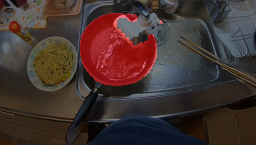} & 
        \includegraphics[width=\linewidth]{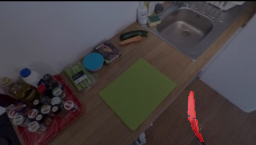} \tabularnewline

         \centering \renewcommand{\familydefault}{\sfdefault}     \fontfamily{phv}\selectfont\fontsize{6.5pt}{8pt}\selectfont\textbf{AffLLM} & 
        \includegraphics[width=\linewidth]{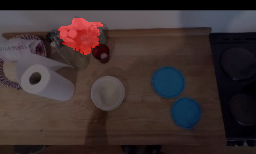} & 
        \includegraphics[width=\linewidth]{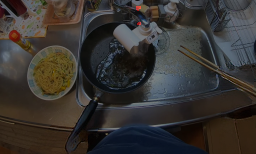} & 
        \includegraphics[width=\linewidth]{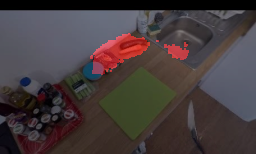} \tabularnewline
        
         \centering \renewcommand{\familydefault}{\sfdefault}     \fontfamily{phv}\selectfont\fontsize{6.5pt}{8pt}\selectfont\textbf{3DOI} & 
        \includegraphics[width=\linewidth]{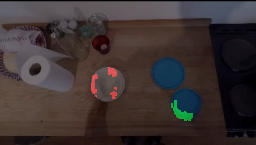} & 
        \includegraphics[width=\linewidth]{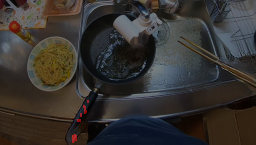} & 
        \includegraphics[width=\linewidth]{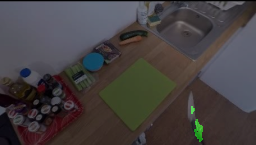} \tabularnewline
        
         \centering \renewcommand{\familydefault}{\sfdefault}     \fontfamily{phv}\selectfont\fontsize{6.5pt}{8pt}\selectfont\textbf{2HAff} & 
        \includegraphics[width=\linewidth]{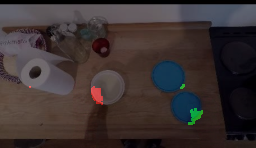} & 
        \includegraphics[width=\linewidth]{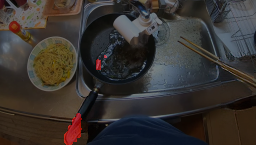} & 
        \includegraphics[width=\linewidth]{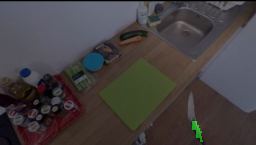} \tabularnewline
        
         \centering \renewcommand{\familydefault}{\sfdefault}     \fontfamily{phv}\selectfont\fontsize{6.5pt}{8pt}\selectfont\textbf{Affordance\\Extraction} & 
        \includegraphics[width=\linewidth]{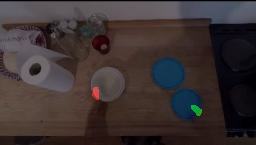} & 
        \includegraphics[width=\linewidth]{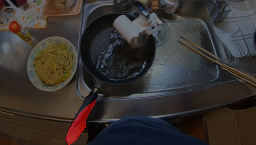} & 
        \includegraphics[width=\linewidth]{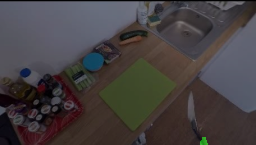} \tabularnewline
        
         \centering \renewcommand{\familydefault}{\sfdefault}     \fontfamily{phv}\selectfont\fontsize{6.5pt}{8pt}\selectfont\textbf{Benchmark\\(Ground Truth)} & 
        \includegraphics[width=\linewidth]{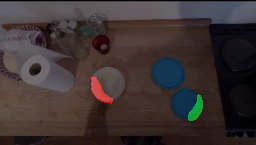} & 
        \includegraphics[width=\linewidth]{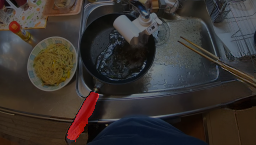} & 
        \includegraphics[width=\linewidth]{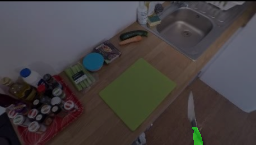} \tabularnewline
    \end{tabular}
    \caption{Qualitative affordance prediction results on the ActAffordance benchmark. We compare our 2HandedAfforder model against LOCATE~\cite{li2023locate}, VRB~\cite{bahl2023affordances}, LISA~\cite{lai2024lisa}, AffordanceLLM~\cite{qian2024affordancellm}, 3DOI~\cite{qian2023understanding}. We also include an example result if we run our affordance extraction method on the activity sequence to show the quality of the extraction. Red and green masks denote left and right hand affordance mask predictions, respectively.}
    \label{fig:image_grid}
    \vspace{-1.5em}
\end{figure}

\begin{table*}[ht!]
    \centering
    \resizebox{\textwidth}{!}{
        \begin{tabular}{|c||c|c|c|c|c||c|c|c|c|c||c|c|c|c|c|}
            \hline
            \multicolumn{16}{|c|}{\textbf{ActAffordance Benchmark}} \\
            \hline
            \textbf{Model} & \multicolumn{5}{c||}{\textbf{EPIC-KITCHENS}} & \multicolumn{5}{c||}{\textbf{EGO4D}} & \multicolumn{5}{c|}{\textbf{Combined}} \\
            & \textbf{IoU $\uparrow$} & \textbf{Precision $\uparrow$} & \textbf{HD $\downarrow$} & \textbf{Dir. HD $\downarrow$} & \textbf{mAP $\uparrow$} & \textbf{IoU $\uparrow$} & \textbf{Precision $\uparrow$} & \textbf{HD $\downarrow$} & \textbf{Dir. HD $\downarrow$} & \textbf{mAP $\uparrow$} & \textbf{IoU $\uparrow$} & \textbf{Precision $\uparrow$} & \textbf{HD $\downarrow$} & \textbf{Dir. HD $\downarrow$} & \textbf{mAP $\uparrow$}\\ 
            \hhline{|=||=|=|=|=|=||=|=|=|=|=||=|=|=|=|=|}
            LISA~\cite{lai2024lisa}        & 0.048 & 0.056 & 298 & 260 & 0.053 & 0.038 & 0.098 & 336 & 257 & 0.084 & 0.044 & 0.050 & 303 & 255 & 0.047 \\
            LOCATE~\cite{li2023locate}      & 0.010 & 0.014 & 274 & 261 & 0.007 & -     & -     & -   & -   & -     & -     & -     & -   & -   & -     \\
            AffLLM~\cite{qian2024affordancellm}      & 0.010 & 0.010 & 267 & 205 & 0.010 & 0.015 & 0.016 & \textbf{229} & \textbf{226} & 0.014 & 0.012 & 0.013 & 287 & 225 & 0.012 \\ 
            2HAffCLIP   & 0.032 & 0.077 & 359 & 317 & 0.068 & 0.023 & 0.050 & 306 & 250 & 0.047 & 0.026 & 0.064 & 341 & 292 & 0.059 \\
            2HAff    & \textbf{0.064} & \textbf{0.125} & \textbf{241} & \textbf{185} & \textbf{0.104} & \textbf{0.051} & \textbf{0.137} & 292 & 227 & \textbf{0.105} & \textbf{0.058} & \textbf{0.130} & \textbf{262} & \textbf{202} & \textbf{0.104} \\
            \hhline{|=||=|=|=|=|=||=|=|=|=|=||=|=|=|=|=|}
            AffExtract  & 0.136 & 0.334 & 199 & 169 & -     & 0.253 & 0.541 & 163 & 121 & -     & 0.185 & 0.420 & 184 & 145 & -     \\
            \hline
        \end{tabular}
    }
    \label{tab:uncropped}
\end{table*}

\vspace{-5pt} 

\begin{table*}[ht!]
    \centering
    \resizebox{\textwidth}{!}{
        \begin{tabular}{|c||c|c|c|c|c||c|c|c|c|c||c|c|c|c|c|}
            \hline
            \multicolumn{16}{|c|}{\textbf{ActAffordance -- Cropped Benchmark}} \\
            \hline
            \textbf{Model} & \multicolumn{5}{c||}{\textbf{EPIC-KITCHENS}} & \multicolumn{5}{c||}{\textbf{EGO4D}} & \multicolumn{5}{c|}{\textbf{Combined}} \\
            & \textbf{IoU $\uparrow$} & \textbf{Precision $\uparrow$} & \textbf{HD $\downarrow$} & \textbf{Dir. HD $\downarrow$} & \textbf{mAP $\uparrow$} & \textbf{IoU $\uparrow$} & \textbf{Precision $\uparrow$} & \textbf{HD $\downarrow$} & \textbf{Dir. HD $\downarrow$} & \textbf{mAP $\uparrow$} & \textbf{IoU $\uparrow$} & \textbf{Precision $\uparrow$} & \textbf{HD $\downarrow$} & \textbf{Dir. HD $\downarrow$} & \textbf{mAP $\uparrow$} \\
            \hhline{|=||=|=|=|=|=||=|=|=|=|=||=|=|=|=|=|}
            LISA~\cite{lai2024lisa}      & \textbf{0.082} & 0.115 & 177 & 111 & 0.110 & 0.097 & 0.132 & 205 & 134 & 0.125 & 0.082 & 0.122 & 196 & 130 & 0.116 \\
            
            LOCATE~\cite{li2023locate}    & 0.026 & 0.097 & 169 & 132 & 0.054 & -     & -     & -   & -   & -     & -     & -     & -   & -   & - \\
            
            AffLLM~\cite{qian2024affordancellm}    & 0.066 & 0.092 & \textbf{155} & \textbf{82}  & 0.088 & 0.091 & 0.139 & \textbf{155} & \textbf{66}  & 0.124   & 0.076 & 0.112 & \textbf{155} & \textbf{76}  & 0.103  \\
            
            VRB~\cite{bahl2023affordances}       & 0.020 & 0.091 & 161 & 152 & -     & 0.018 & 0.083 & 175 & 160 & -     & 0.019 & 0.088 & 167 & 155 & - \\

            3DOI~\cite{qian2023understanding} & 0.038 & \textbf{0.227} & 337 & 289 & 0.188 & 0.071 & 0.221 & 182 & 110 & 0.168 & 0.082 & 0.224 & 168 & 109 & 0.180 \\
            2HAffCLIP & 0.038 & 0.144 & 170 & 108 & 0.131 & 0.040 & 0.202 & 176 & 98  & 0.186 & 0.039 & 0.168 & 172 & 104 & 0.154 \\
            
            2HAff  & 0.074 & 0.223 & 188 & 114 & \textbf{0.204} & \textbf{0.101} & \textbf{0.331} & 169 & 80  & \textbf{0.291} & \textbf{0.086} & \textbf{0.269} & 180 & 100 & \textbf{0.240} \\
            
            \hline
        \end{tabular}
    }
    \caption{Comparison of our models and baseline methods on the ActAffordance Benchmark (top) and the modified version ActAffordance-Cropped (bottom) where images are cropped to a bounding-box around the target objects. Performance is evaluated separately on the EPIC-KITCHENS and EGO4D splits, as well as on the combined benchmark. The reported metrics include IoU (Intersection over Union), Precision, HD (Hausdorff Distance), Dir. HD (Directional Hausdorff Distance), and mAP (Mean Average Precision). For mAP, we average over five different thresholds, and the values for the other metrics correspond to the highest scores obtained across these thresholds. We also run our affordance extraction method, AffExtract, on the activity sequences in the benchmark as a measure of data quality and alignment with the benchmark annotations.}
    \label{tab:cropped}
    \vspace{-1.5em}
\end{table*}

\subsection{Metrics for Evaluation}
Since we treat the affordance detection problem as a segmentation task, we use the following metrics to evaluate the performance of the proposed models and baselines: precision, intersection over union (IoU) and the directed and general Hausdorff distance (HD). We train our 2HandedAfforder and 2HandedAfforder-CLIP models on the 2HANDS dataset and evaluate on the ``ActAffordance" benchmark. We evaluate performance on both the EPIC-KITCHENS and Ego4D splits of the benchmark. Note that there is no overlap between the data from EPIC-KITCHENS used in 2HANDS. The evaluation on the Ego4D split of the benchmark also helps answer the generalization question since there is no Ego4D data in 2HANDS.

Note that for the evaluation of our models, false negative predictions play a reduced role since our models are not trained to predict all multimodal solutions in the benchmark but to predict \textit{precise} affordance regions which might only cover a subset of the possible solutions. Thus, the key metric for comparison is \textbf{precision} over IoU. Another common segmentation metric is Hausdorff distance (HD). For each point in each set, the distance to the closest point from the other set is computed and the Hausdorff distance is the maximum of all of these distances. Similar to the IoU case, including the distance from ground truth to the prediction might distort the results since we aim to predict precise affordances that may only cover a smaller subset of the ground truth. Thus, we also provide the directed Hausdorff distance that only calculates the maximum distance from the prediction set points to the ground truth set.

To further show the applicability of our approach to real world robotics scenarios, we evaluate our model in-the-wild in a kitchen environment on various household objects. To show that our model can provide useful actionable affordances, we test the predictions on a real-robot system in this kitchen environment. Specifically, we use an RGBD camera mounted on a mobile manipulator robot and use the affordances predicted by our model to segment RGB images and obtain segmented point clouds. These segmented pointclouds denote where the robot should grasp objects to perform a manipulation task. For manipulation, we use pre-designed manipulation primitives for the robot and perform grasping using a 6DoF grasp prediction network.

\begin{figure*}[t]
    \centering
    
    \begin{subfigure}[b]{0.195\textwidth}
           \centering \renewcommand{\familydefault}{\sfdefault}     \fontfamily{phv}\selectfont
        \caption*{\textbf{stir vegetables}}
    \end{subfigure}
    \hfill
    \begin{subfigure}[b]{0.195\textwidth}
           \centering \renewcommand{\familydefault}{\sfdefault}     \fontfamily{phv}\selectfont
        \caption*{\textbf{pour into cup}}
    \end{subfigure}
    \hfill
    \begin{subfigure}[b]{0.195\textwidth}
           \centering \renewcommand{\familydefault}{\sfdefault}     \fontfamily{phv}\selectfont
        \caption*{\textbf{pick up pot}}
    \end{subfigure}
    \hfill
    \begin{subfigure}[b]{0.195\textwidth}
           \centering \renewcommand{\familydefault}{\sfdefault}     \fontfamily{phv}\selectfont
        \caption*{\textbf{open pot}}
    \end{subfigure}
    \hfill
    \begin{subfigure}[b]{0.195\textwidth}
           \centering \renewcommand{\familydefault}{\sfdefault}     \fontfamily{phv}\selectfont
        \caption*{\textbf{open bottle}}
    \end{subfigure}
    
    
    \begin{subfigure}[b]{0.195\textwidth}
        \centering
        \includegraphics[width=\textwidth, trim={0 0 1.2635cm 1cm},clip]{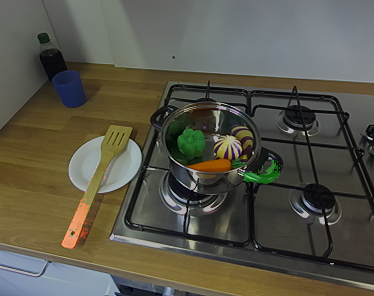}
    \end{subfigure}
    \hfill
    \begin{subfigure}[b]{0.195\textwidth}
        \centering
        \includegraphics[width=\textwidth]{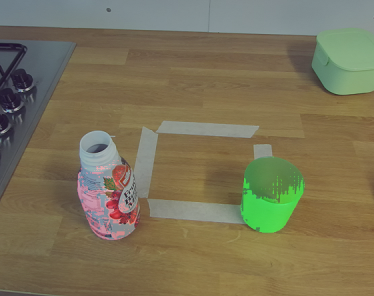}
    \end{subfigure}
    \hfill
    \begin{subfigure}[b]{0.195\textwidth}
        \centering
        \includegraphics[width=\textwidth]{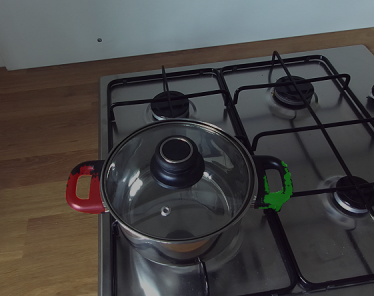}
    \end{subfigure}
    \hfill
    \begin{subfigure}[b]{0.195\textwidth}
        \centering
        \includegraphics[width=\textwidth]{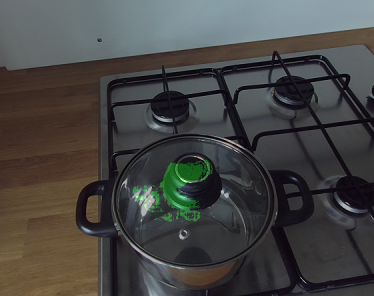}
    \end{subfigure}
    \hfill
    \begin{subfigure}[b]{0.195\textwidth}
        \centering
        \includegraphics[width=\textwidth]{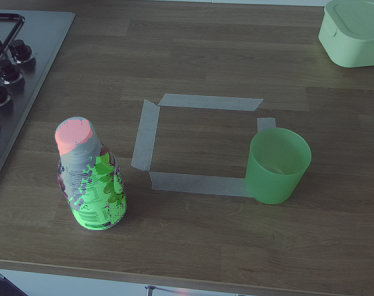}
    \end{subfigure}
    
    
    \begin{subfigure}[b]{0.195\textwidth}
        \centering
        \includegraphics[width=\textwidth, trim={0 5cm 6.79cm 0},clip]{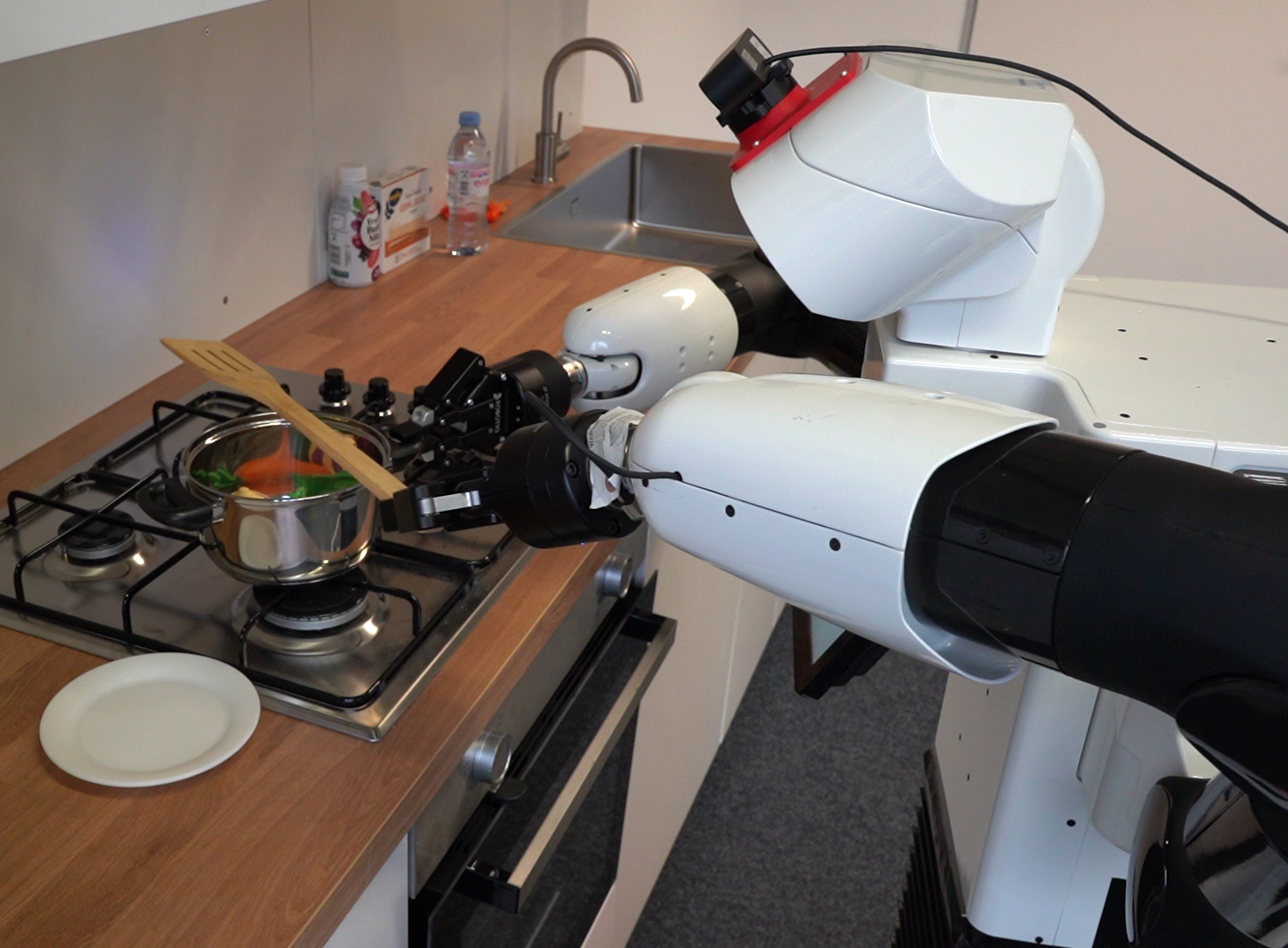}
    \end{subfigure}
    \hfill
    \begin{subfigure}[b]{0.195\textwidth}
        \centering
        \includegraphics[width=\textwidth, trim={0 5cm 6.79cm 0},clip]{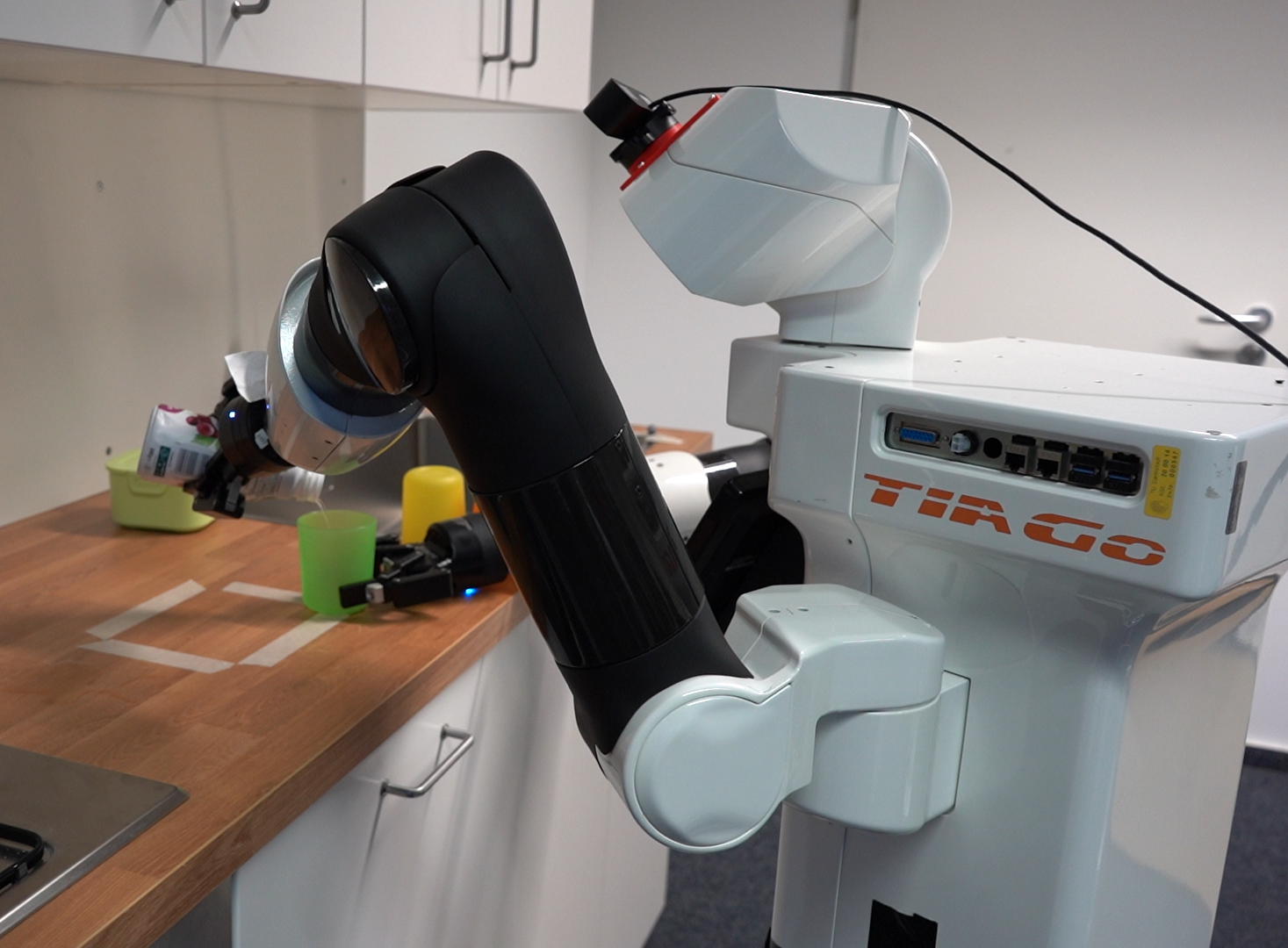}
    \end{subfigure}
    \hfill
    \begin{subfigure}[b]{0.195\textwidth}
        \centering
        \includegraphics[width=\textwidth, trim={0 5cm 6.79cm 0},clip]{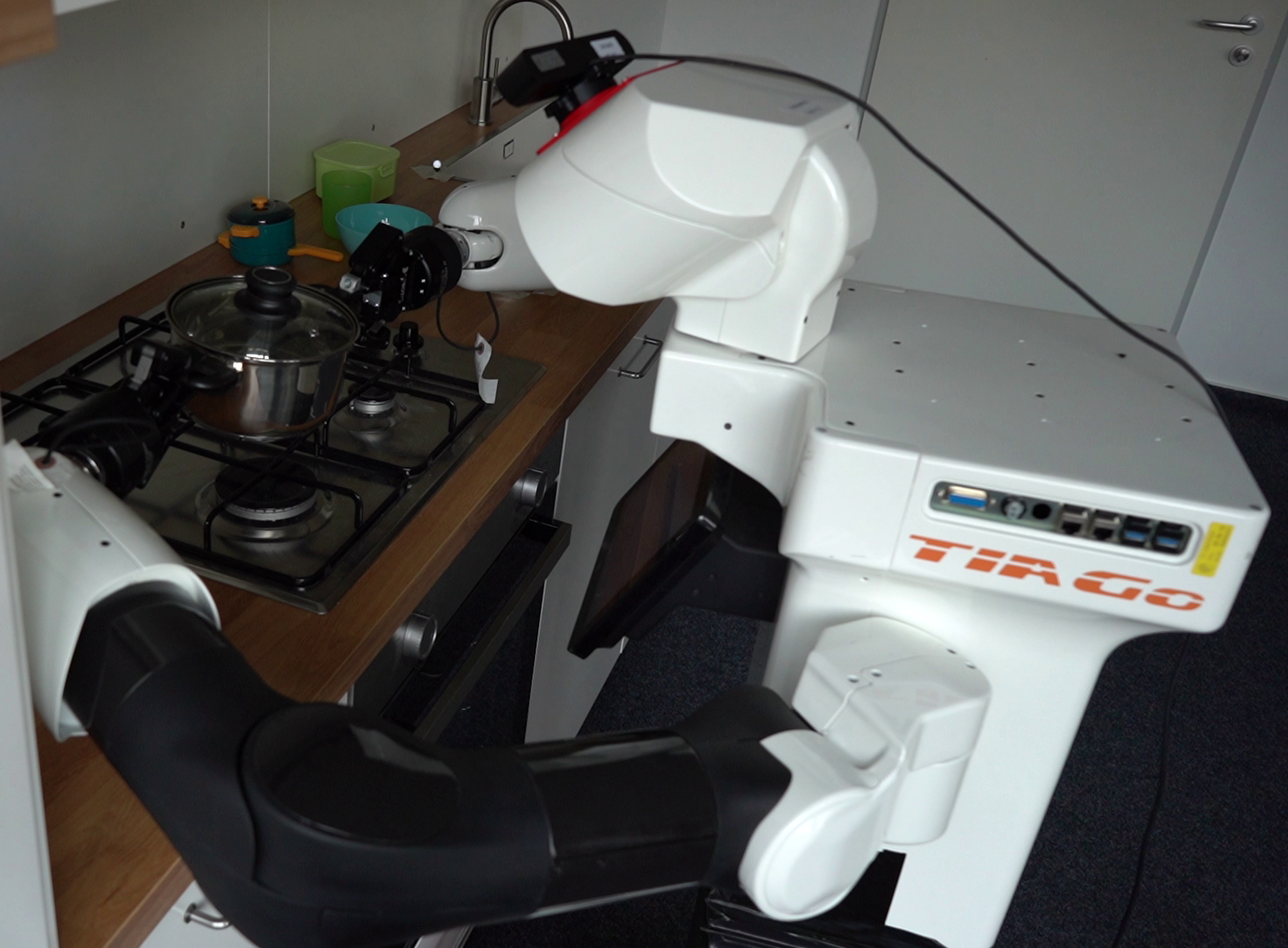}
    \end{subfigure}
    \hfill
    \begin{subfigure}[b]{0.195\textwidth}
        \centering
        \includegraphics[width=\textwidth, trim={0 5cm 6.79cm 0},clip]{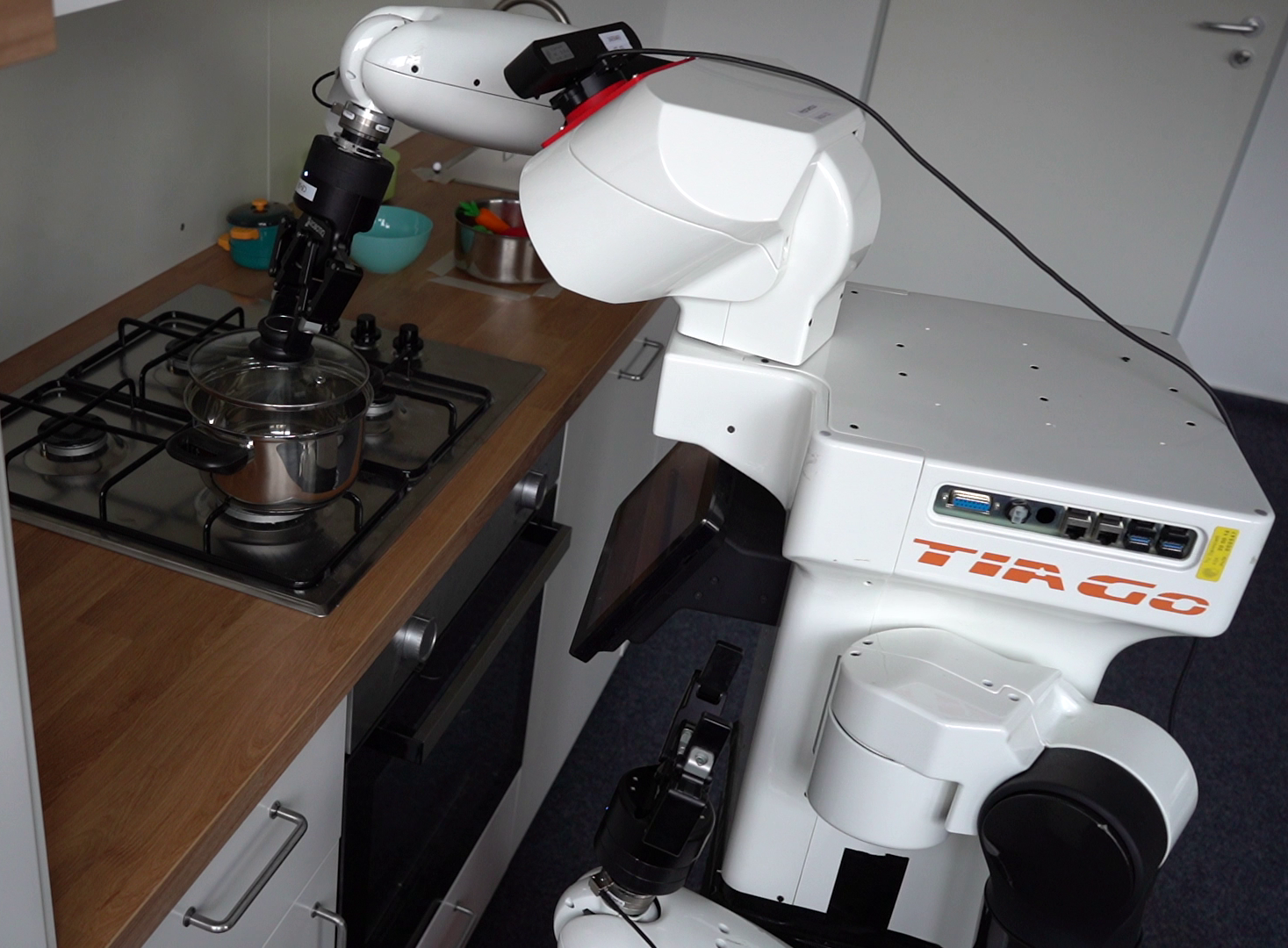}
    \end{subfigure}
    \hfill
    \begin{subfigure}[b]{0.195\textwidth}
        \centering
        \includegraphics[width=\textwidth, trim={0 5cm 6.79cm 0},clip]{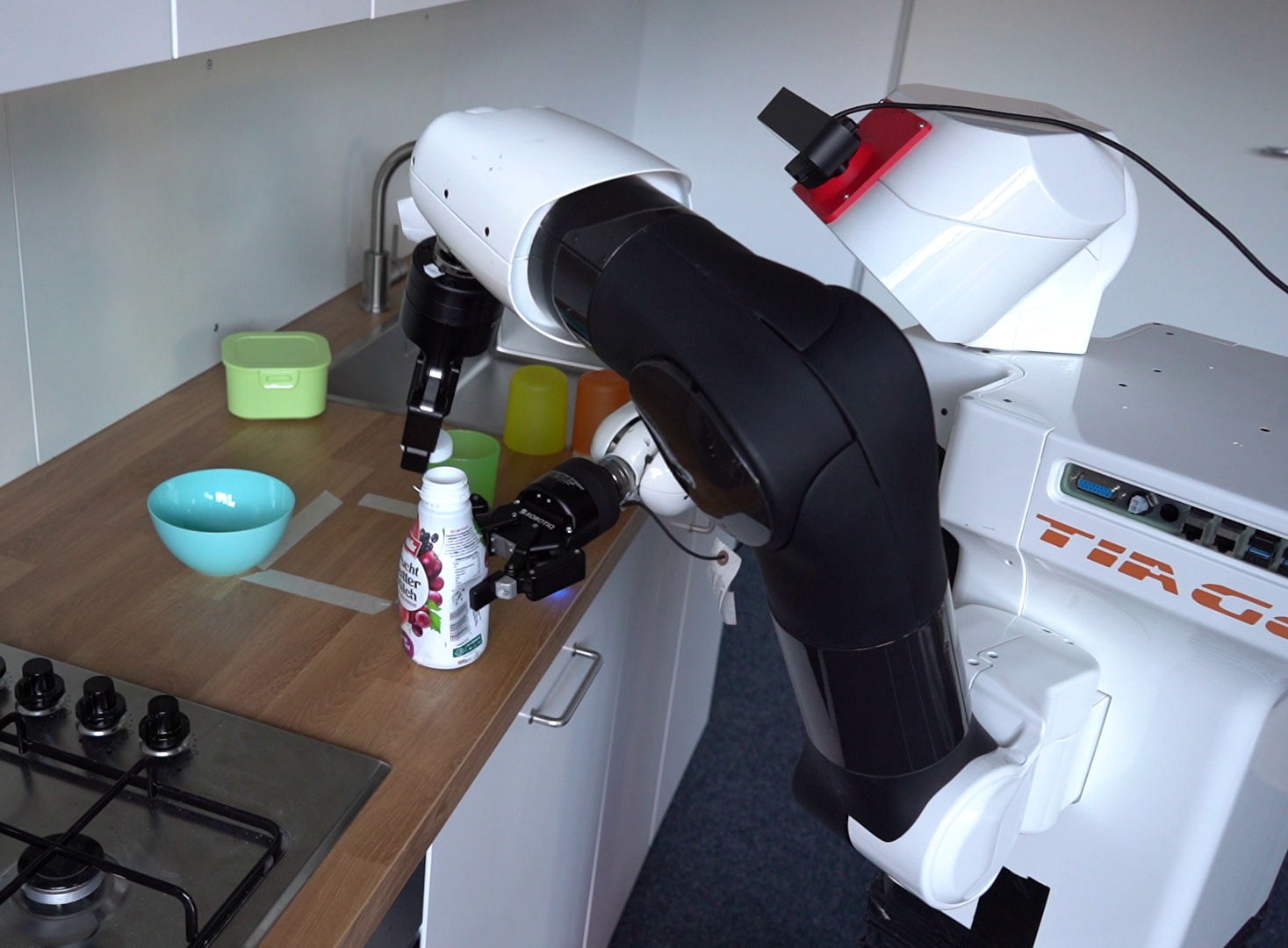}
    \end{subfigure}
    
    \caption{Examples of different manipulation tasks executed on a bimanual Tiago++ robot. Red and green masks denote left and right hand affordance mask predictions, respectively. We segment the task-specific object affordance regions, propose grasps for these regions, and use pre-designed motion primitives to execute manipulation tasks. 
    Videos are available at \href{https://sites.google.com/view/2handedafforder}{sites.google.com/view/2handedafforder}.
    }
    \label{fig:robot_preds}
    \vspace{-1.5em}
\end{figure*}

\section{Results}

\subsection{Affordance Extraction Quality}

We assess the quality of affordances obtained from our extraction pipeline (Sec.~\ref{ssec:affordance-extraction}) by evaluating their alignment with the human annotations in the ActAffordance benchmark. The results are shown in Table~\ref{tab:cropped}, ``AffExtract", and Figure~\ref{fig:image_grid}. As noted before, the benchmark annotations contain all the possible modes of object interaction, while the 
extraction process and our models only cover a single interaction mode. Thus, precision is the most important metric to evaluate over IoU. The same principle is true for the Hausdorff distance (HD), which is why we also report directional Hausdorff distance (Dir. HD), which only calculates the maximum distance from the prediction set points to the ground truth set. 
We note the precision of AffExtract is better for the Ego4D split~(0.541) than the EPIC-KITCHENS split~(0.334) with a combined score of 0.42. This shows a reasonably good alignment with the human-annotated segmentations from the benchmark and meaningful affordance region extraction. The IoU scores are relatively lower, with an average of 0.185, showing the challenge of the task when compared against human-level object understanding.

\subsection{Comparison against baselines on ActAffordance benchmark}

Since ours is the first method to perform bimanual affordance mask detection using text prompts, there exist no directly comparable baselines. Thus, we adapt affordance detection baselines which includes a SOTA text-based reasoning segmentation baseline. 
Since several weakly-supervised affordance detection methods~\cite{goyal2022human, bahl2023affordances, qian2024affordancellm} represent affordances as only points or points+probabalistic heatmaps around them, we adapt their predictions into segmentation masks by choosing different probability thresholds at which pixels are considered to be part of the affordance region. We use the following baselines for comparison: (i)~\textbf{LISA~\cite{lai2024lisa}}, an object segmentation VLM with text-based reasoning capabilities. 
(ii)~\textbf{LOCATE~\cite{li2023locate}} and (iii)~\textbf{AffordanceLLM~\cite{qian2024affordancellm}}, which are trained on explicit affordance labels from the AGD20K dataset~\cite{luo2022learning}. 
(iv)~\textbf{3DOI~\cite{qian2023understanding}}, a fully-supervised method using point-based affordance data from exo and egocentric images and uses query points during inference. 
(v)~\textbf{VRB~\cite{bahl2023affordances}}, which uses bounding boxes to predict affordance hotspots.

All models are evaluated on the ActAffordance benchmark. Additionally, we assess the methods on a modified version of the benchmark, where all images were cropped to encompass the target objects for comparison with VRB that utilizes bounding boxes~\cite{bahl2023affordances} and 3DOI that uses query points~\cite{qian2023understanding} as prompts instead of language. Although no baseline can be trained on our 2HANDS dataset, we make several adjustments to allow inference on the benchmark. For LISA, LOCATE, AffLLM, we ignore left/right classification and compare predicted masks to the union of left and right masks in the benchmark. For VRB, 3DOI, we input the necessary ground-truth bounding boxes and object mask centers (cropped benchmark) and predict separate left/right masks. Since LOCATE~\cite{li2023locate} uses an explicit affordance class label as input, we adapt the EPIC-VISOR verb categories used in 2HANDS to fit the AGD20K classes used in LOCATE. Such an adaptation is not possible for Ego4D so we exclude LOCATE from the comparison on the Ego4D split. To isolate the effect of the 2HANDS dataset, the comparison with AffLLM and LISA is key since their network architecture is close to ours.

Figure~\ref{fig:image_grid} shows some qualitative affordance prediction results and Table~\ref{tab:cropped} shows the quantitative results. On the combined ActAffordance benchmark, 2HandedAfforder achieves the best results across all metrics. LISA is the next best method since it accurately segments the correct object in the scene, resulting in a natural overlap with the ground truth. This demonstrates the power of reasoning segmentation for the challenging task of prompt-based affordance prediction. This reasoning ability is also validated by the 2HandedAfforder-CLIP version being only third-best. Though our models were not trained on any Ego4D data, their performance on Ego4D is still reasonable and often better than the EPIC-KITCHENS split. The IoU scores are low across the board for all methods, indicating further room for improvement on this challenging task.


The results on the cropped version of the benchmark, Table~\ref{tab:cropped} (lower), show similar results with performance improvements across the board since the uncropped benchmark is more difficult. 
In this setting, the other baseline models that use prompts or query points as input can be compared as well. 2HandedAfforder again achieves the best performance on the combined benchmark, with significantly better precision and mAP scores than the uncropped benchmark. 3DOI also performs reasonably in terms of precision. Surprisingly, AffordanceLLM achieves good scores in HD and Dir. HD, even though the IoU scores are lower. This stems from the fact that AffordanceLLM is relatively more optimistic and always predicts some affordance regions. The other methods can sometimes not detect any affordance regions and have no mask predictions, which penalizes the HD and dir. HD significantly. LISA is still the third or fourth best method on most metrics, while VRB, being a task-agnostic method, performs poorly.

\subsection{In-the-wild Affordance Prediction and Robot Demonstration}



We conduct robotic manipulation experiments with various objects using a bimanual Tiago++ robot in a realistic kitchen environment. 
We deploy our 2HandedAfforder model for affordance region segmentation inference based on task prompts such as `pour into cup'.

To enhance the model's performance for real-world application, we obtain object bounding boxes and masks using a prompt-based segmentation method, LangSAM~\cite{langsegmentanything}. We then performed inference on the cropped object images. Moreover, to enhance the stability of our predictions, we only considered the intersection between our inferred affordance masks and the object masks generated by LangSAM. This also allowed us to adjust the prediction threshold to be more optimistic and generate larger affordance masks.


We demonstrate how our affordance prediction method improves the performance of a robot in executing manipulation tasks compared to using standard object or part segmentation approaches, such as the mask output of LangSAM. By integrating our affordance prediction into the grasping pipeline, the robot is able to make more informed grasping decisions, leading to greater task success. Examples of different manipulation tasks are shown in Figure~\ref{fig:robot_preds} and in videos at \href{https://sites.google.com/view/2handedafforder}{sites.google.com/view/2handedafforder}.

\section{Conclusion}
\label{sec:conclusion}

In this work, we proposed a framework for extracting precise, meaningful affordance regions from human activity videos, resulting in the 2HANDS dataset of actionable bimanual affordances. 
We further introduced a novel VLM-based task-aware bimanual affordance prediction model, 2HandedAfforder, that predicts actionable affordance regions from task-related text prompts.
To evaluate the alignment of the extracted affordances with human-annotated ones, we further proposed a novel ActAffordance benchmark, which is a particularly challenging benchmark for prompt-based segmentation of precise object affordance regions.
Our experiments demonstrate that 2HandedAfforder can predict meaningful task-oriented bimanual affordances compared to other works, thereby showcasing the effectiveness of our data extraction pipeline and proposed model.


{
    \small
    \bibliographystyle{ieeenat_fullname}
    \bibliography{main}
}

\clearpage
\setcounter{page}{1}
\maketitlesupplementary

\section{Additional Filtering and Augmentation Steps}
Between each of the major steps of the affordance extraction pipeline different filtering steps were applied to clean up the data. After calculating the intersection of the completed mask and the hand mask erosion and dilation steps were applied to remove scattered mask pixels and fill gaps to leave only one connected affordance mask. Furthermore, inconsistencies and inaccuracies within the data were detected and the data points were deleted, e.g. the calculated affordance masks are empty, the action is classified as bimanual but only one affordance mask is provided. Lastly, we remove datapoints with narrations that are too vague or do not describe an affordance by blacklisting some expressions, e.g. `throw \textbf{something} into the bin' or `\textbf{look} at pan'. We augment the data by flipping all of the actions horizontally, essentially doubling the size of the dataset. By doing that we even out the ratio of left-handed and right-handed actions. Afterwards, we apply common augmenting strategies also used by Goyal et al. \cite{goyal2022human}, i.e., color jittering (randomly changing the brightness, contrast, saturation and hue of the inpainted frame) and cropping.


\section{2HANDS Dataset}
Each data point of 2HANDS consists of an inpainted frame, two affordance masks where one of them is left empty if it is a unimanual action and the narration. We also provide additional information such as the object masks and object names if needed. In the end, the proposed dataset 2HANDS consists of over 278k datapoints from 25 different kitchen environments from the EPIC-KITCHENS dataset. An overview of the dataset can be found in Table 4.1. This dataset was used to train the models.
\begin{table}[ht]
\centering
\begin{tabular}{|>{\bfseries}l|r|}
\hline
 & \textbf{Amount} \\
\hline
\textbf{Left Handed} & 76,278 \\
\hline
\textbf{Right Handed} & 76,278 \\
\hline
\textbf{Symmetric} & 51,684 \\
\hline
\textbf{Asymmetric} & 73,920 \\
\hline
\textbf{Total} & 278,160 \\
\hline
\textbf{No. Kitchen Environments} & 25 \\
\hline
\textbf{No. Videos} & 47 \\
\hline
\textbf{No. Object Classes} & 160 \\
\hline
\textbf{No. Verb Classes} & 73 \\
\hline
\end{tabular}
\caption{Overview of the dataset}
\end{table}

We collected affordance masks for 160 different object categories and 73 verb class.
\\

The object classes: \\
alarm, almond, aubergine, bag, banana, basil, bean:green, beer, bin, board:chopping, book, bottle, bowl, box, bread, broccoli, brush, butter, button, can, candle, cap, caper, carrot, chair, cheese, cherry, chicken, chilli, choi:pak, chopstick, cinnamon, cloth, clothes, coffee, colander, container, cooker:slow, cork, corn, cover, cucumber, cumin, cup, cupboard, cutlery, cutter:pizza, dishwasher, dough, drawer, fan:extractor, filter, fish, flour, food, fork, fridge, garlic, ginger, glass, glove, grater, hand, heat, heater, hob, holder, ice, jar, jug, juicer, kettle, knife, knob, label, ladle, leaf, leek, lemon, lettuce, lid, light, lighter, liquid:washing, machine:sous:vide, machine:washing, maker:coffee, mat, meat, microwave, milk, mixture, mushroom, napkin, noodle, nut, oil, onion, opener:bottle, oven, package, pan, pan:dust, paper, paste, peach, peeler:potato, pepper, phone, pin:rolling, pith, pizza, plate, plug, pork, pot, potato, powder:washing, processor:food, rack:drying, rest, rice, roll, rubbish, salt, sauce, sausage, scissors, shell:egg, sink, skin, soda, spatula, sponge, spoon, sprout, stalk, stock, syrup, tap, toaster, tofu, tomato, tongs, top, towel, towel:kitchen, tray, utensil, vegetable, vinegar, wall, water, whetstone, window, wine, wire, wrap, wrap:plastic, yoghurt \\

The verb classes: \\
add, adjust, apply, attach, break, brush, carry, check, choose, close, coat, cook, crush, cut, divide, drink, dry, empty, fill, filter, flatten, flip, form, gather, hold, increase, insert, knead, lift, lower, mix, move, open, pat, peel, pour, press, pull, put, remove, rip, roll, rub, scoop, scrape, screw, scrub, season, serve, set, shake, sharpen, slide, soak, sort, spray, sprinkle, squeeze, stab, stretch, take, throw, turn, turn-down, turn-off, turn-on, uncover, unroll, unscrew, unwrap, use, wash, wrap\\

\textit{The full dataset and codebase will be released at \href{https://sites.google.com/view/2handedafforder}{sites.google.com/view/2handedafforder}.}\\

\section{Additional Qualitative Results}
\begin{figure*}[h!]
    \centering
    \renewcommand{\arraystretch}{1.0} 
    \setlength{\tabcolsep}{1pt} 
    
    \begin{tabular}{@{}m{0.45\textwidth}@{\hskip 1pt}m{0.45\textwidth}@{}}
    
        \centering \renewcommand{\familydefault}{\sfdefault}     \fontfamily{phv}\selectfont\fontsize{6.5pt}{8pt}\selectfont\textbf{take onion} &  
        \centering \renewcommand{\familydefault}{\sfdefault}     \fontfamily{phv}\selectfont\fontsize{6.5pt}{8pt}\selectfont\textbf{pick up bowl}
        \tabularnewline
        
         \centering \renewcommand{\familydefault}{\sfdefault}   
        \includegraphics[width=\linewidth]{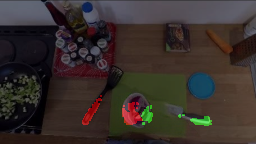} & 
        \includegraphics[width=\linewidth]{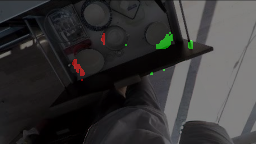}
    \end{tabular}
    \caption{Failure cases. In both cases the model is undecisive of what to do. In the left example it predicts affordance regions at different objects that are also not related to the task, i.e., the spatula and the knife. In the right example the model predicts a bimanual action to pick up the bowl and predicts affordance regions at multiple bowls even though only one bowl is supposed to be picked up.}
    \label{fig:image_grid2}
\end{figure*}

\begin{figure*}[h!]
    \centering
    \renewcommand{\arraystretch}{1.0} 
    \setlength{\tabcolsep}{1pt} 
    
    \begin{tabular}{@{}m{0.12\textwidth}@{\hskip 1pt}m{0.16\textwidth}@{\hskip 1pt}m{0.16\textwidth}@{\hskip 1pt}m{0.16\textwidth}@{\hskip 1pt}m{0.16\textwidth}@{\hskip 1pt}m{0.16\textwidth}@{}}
    
        &  \centering \renewcommand{\familydefault}{\sfdefault}     \fontfamily{phv}\selectfont\fontsize{6.5pt}{8pt}\selectfont\textbf{wash the cooking pot} &  
        \centering \renewcommand{\familydefault}{\sfdefault}     \fontfamily{phv}\selectfont\fontsize{6.5pt}{8pt}\selectfont\textbf{take knife} &  
        \centering \renewcommand{\familydefault}{\sfdefault}     \fontfamily{phv}\selectfont\fontsize{6.5pt}{8pt}\selectfont\textbf{close lid} & 
        \centering \renewcommand{\familydefault}{\sfdefault}     \fontfamily{phv}\selectfont\fontsize{6.5pt}{8pt}\selectfont\textbf{put down plate} &
        \centering \renewcommand{\familydefault}{\sfdefault}     \fontfamily{phv}\selectfont\fontsize{6.5pt}{8pt}\selectfont\textbf{put down bowl}
        \tabularnewline
        
         \centering \renewcommand{\familydefault}{\sfdefault}     \fontfamily{phv}\selectfont\fontsize{6.5pt}{8pt}\selectfont\textbf{LISA} & 
        \includegraphics[width=\linewidth]{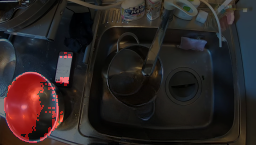} & 
        \includegraphics[width=\linewidth]{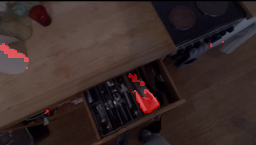} & 
        \includegraphics[width=\linewidth]{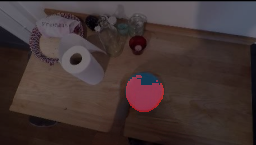} &
        \includegraphics[width=\linewidth]{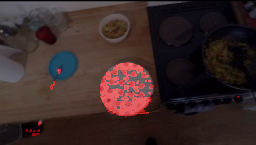} &
        \includegraphics[width=\linewidth]{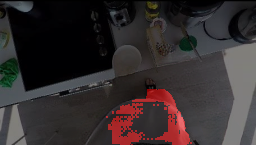} \tabularnewline

         \centering \renewcommand{\familydefault}{\sfdefault}     \fontfamily{phv}\selectfont\fontsize{6.5pt}{8pt}\selectfont\textbf{AffLLM} & 
        \includegraphics[width=\linewidth]{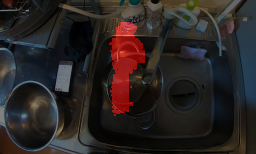} & 
        \includegraphics[width=\linewidth]{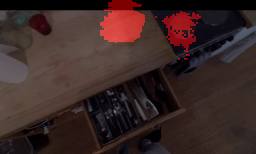} & 
        \includegraphics[width=\linewidth]{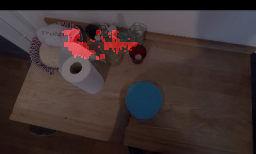} & 
        \includegraphics[width=\linewidth]{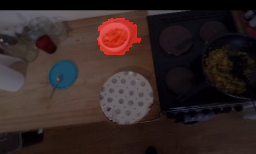} &
        \includegraphics[width=\linewidth]{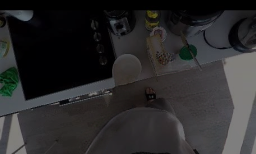} \tabularnewline
        
         \centering \renewcommand{\familydefault}{\sfdefault}     \fontfamily{phv}\selectfont\fontsize{6.5pt}{8pt}\selectfont\textbf{3DOI} & 
        \includegraphics[width=\linewidth]{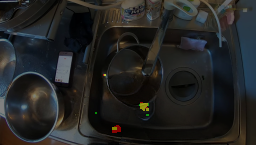} & 
        \includegraphics[width=\linewidth]{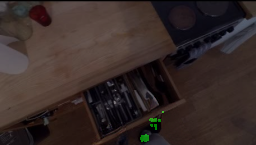} & 
        \includegraphics[width=\linewidth]{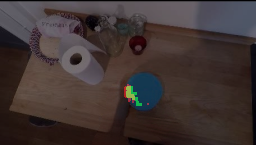} & 
        \includegraphics[width=\linewidth]{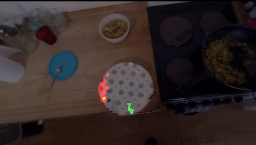} & 
        \includegraphics[width=\linewidth]{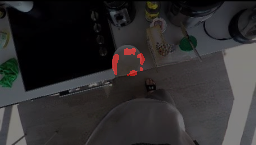} \tabularnewline
        
         \centering \renewcommand{\familydefault}{\sfdefault}     \fontfamily{phv}\selectfont\fontsize{6.5pt}{8pt}\selectfont\textbf{2HAff} & 
        \includegraphics[width=\linewidth]{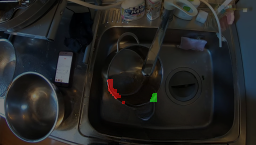} & 
        \includegraphics[width=\linewidth]{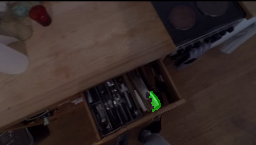} & 
        \includegraphics[width=\linewidth]{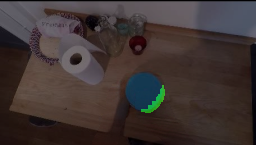} & 
        \includegraphics[width=\linewidth]{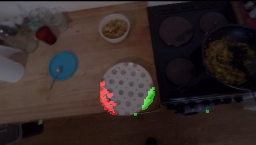} &
        \includegraphics[width=\linewidth]{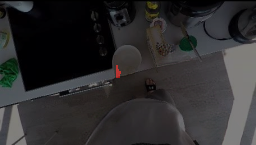} \tabularnewline
        
         \centering \renewcommand{\familydefault}{\sfdefault}     \fontfamily{phv}\selectfont\fontsize{6.5pt}{8pt}\selectfont\textbf{Affordance\\Extraction} & 
        \includegraphics[width=\linewidth]{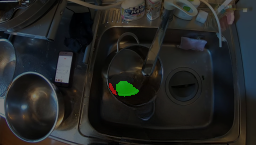} & 
        \includegraphics[width=\linewidth]{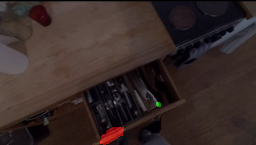} & 
        \includegraphics[width=\linewidth]{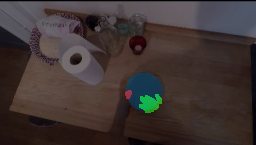} & 
        \includegraphics[width=\linewidth]{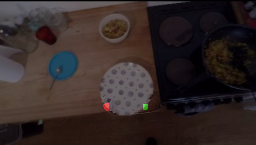} & 
        \includegraphics[width=\linewidth]{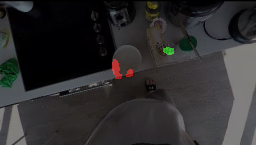} \tabularnewline
        
         \centering \renewcommand{\familydefault}{\sfdefault}     \fontfamily{phv}\selectfont\fontsize{6.5pt}{8pt}\selectfont\textbf{Benchmark\\(Ground Truth)} & 
        \includegraphics[width=\linewidth]{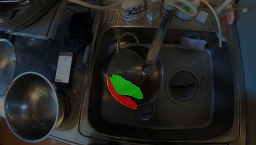} & 
        \includegraphics[width=\linewidth]{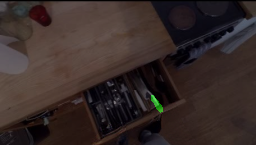} & 
        \includegraphics[width=\linewidth]{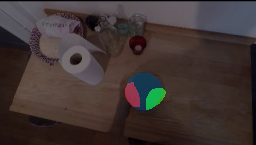} & 
        \includegraphics[width=\linewidth]{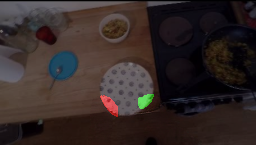} & 
        \includegraphics[width=\linewidth]{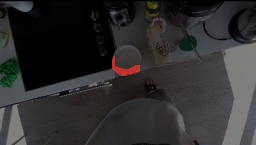} \tabularnewline
    \end{tabular}
    \caption{Additional qualitative results showing the performance of our proposed model compared to different baselines and the ground truth.}
    \label{fig:image_grid3}
\end{figure*}
\newpage
\section{ActAffordance Annotation Procedure}
\label{sec:actaffordance-annotation}
For annotating the images for the ActAffordance Benchmark, we used TORAS~\cite{torontoannotsuite}. We asked 10 human annotators to highlight all possible interaction regions of the target objects in the image where the hands were already removed with respect to the underlying task, i.e. the narration. This annotation was done for both the left and right hands. Additionally, annotators also had access to the original image to see how the hands interacted with the objects in the scene.

\section{Real robot experiments}

\begin{figure}[h]
    \centering
    \begin{subfigure}[b]{0.45\textwidth}
        \centering
        \includegraphics[width=\textwidth]{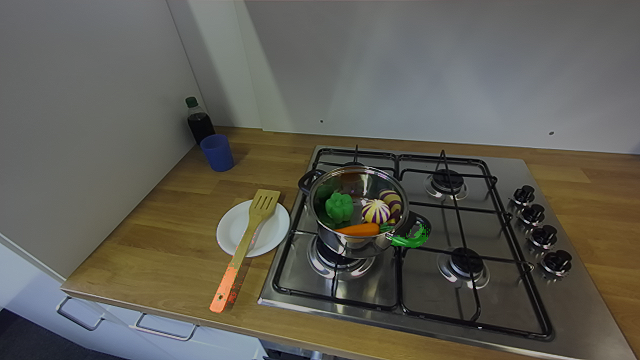}
    \end{subfigure}
    \hfill
    \begin{subfigure}[b]{0.45\textwidth}
        \centering
        \includegraphics[width=\textwidth]{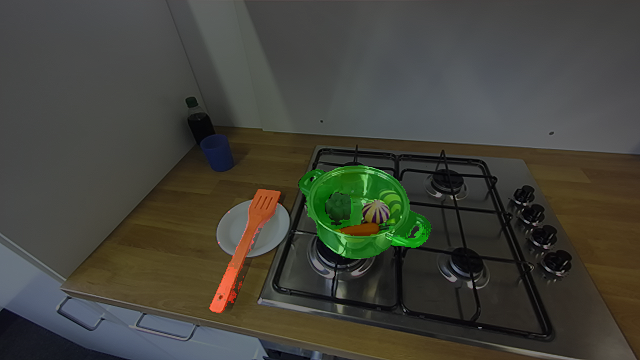}
    \end{subfigure}
    \caption{The affordance detection of our method detects precise affordance regions (left) for the pot and the spatula that can be used to successfully perform the task of stirring within the pot. The right image shows the mask outputs by LangSAM. The text prompts used for this prediction were ``wooden spatula" and ``cooking pot" for LangSAM and ``stir vegetables" for 2HandedAfforder.}
    \label{fig:robot_example}
\end{figure}

A successful example for an affordance prediction as well as the corresponding masks from LangSAM are visualized in Figure~\ref{fig:robot_example}. \\

\begin{figure}[h]
    \centering
    \includegraphics[width=0.6\columnwidth]{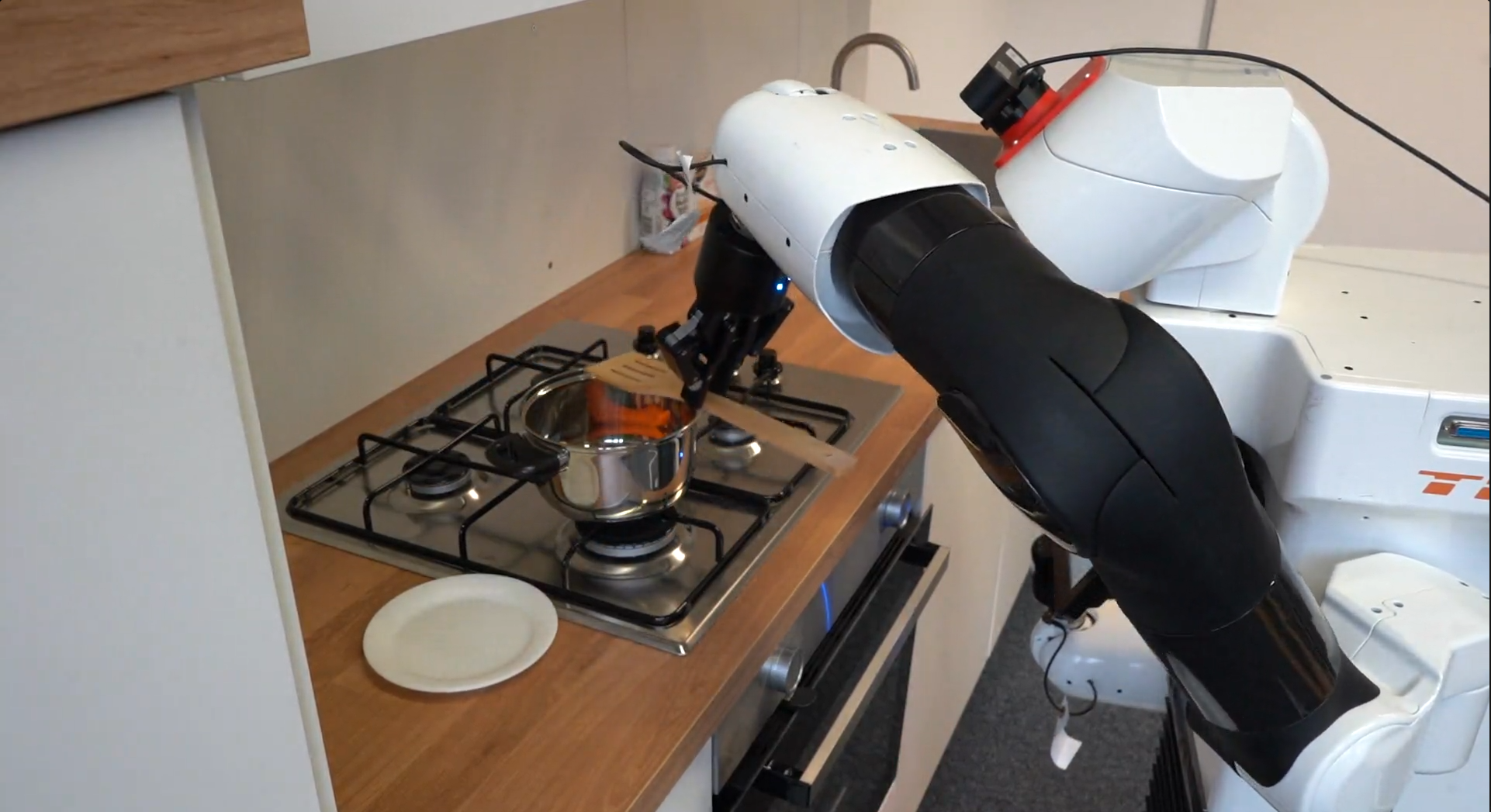}
    
    \includegraphics[width=0.6\columnwidth]{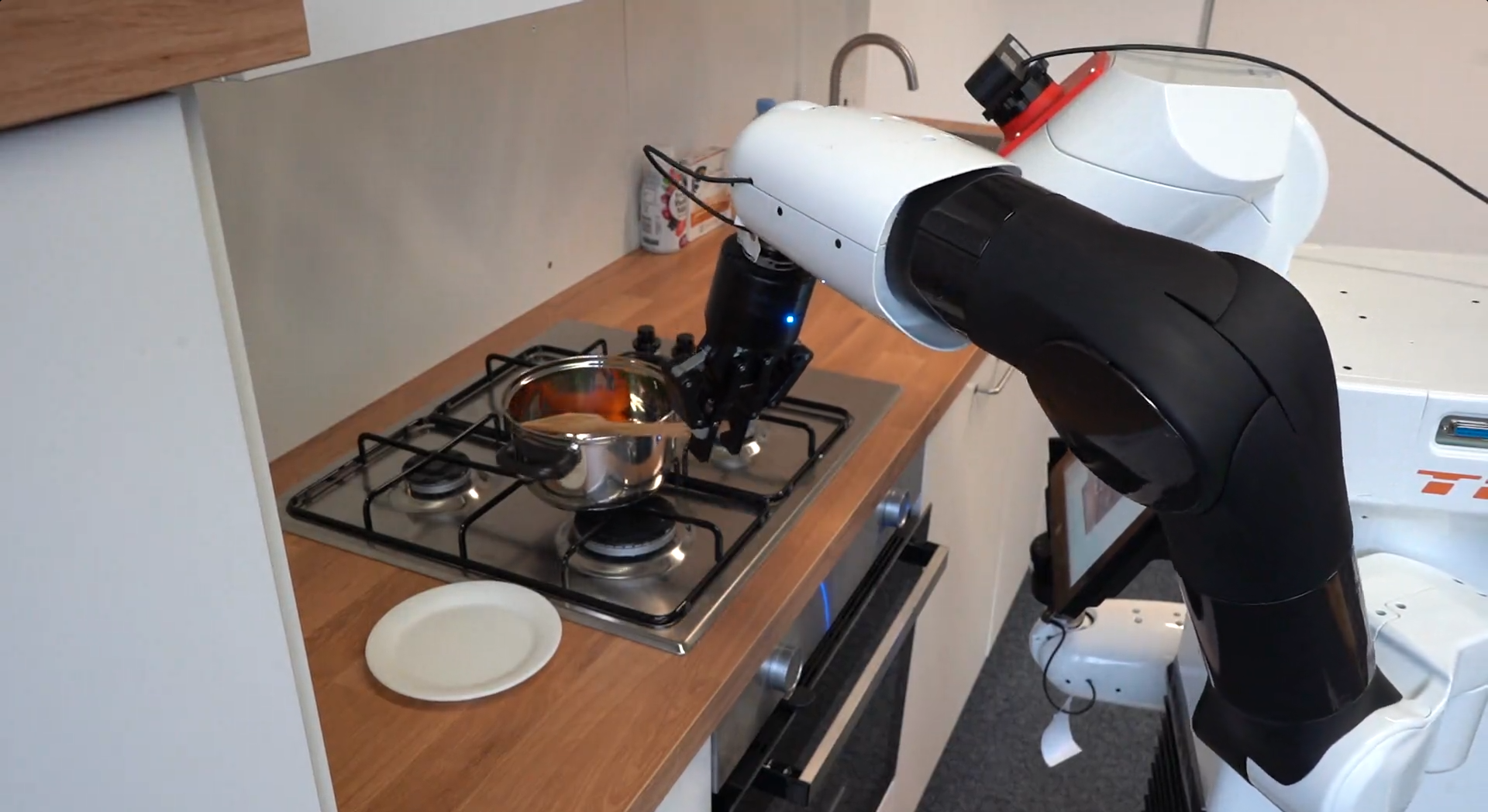}
    
    \includegraphics[width=0.6\columnwidth]{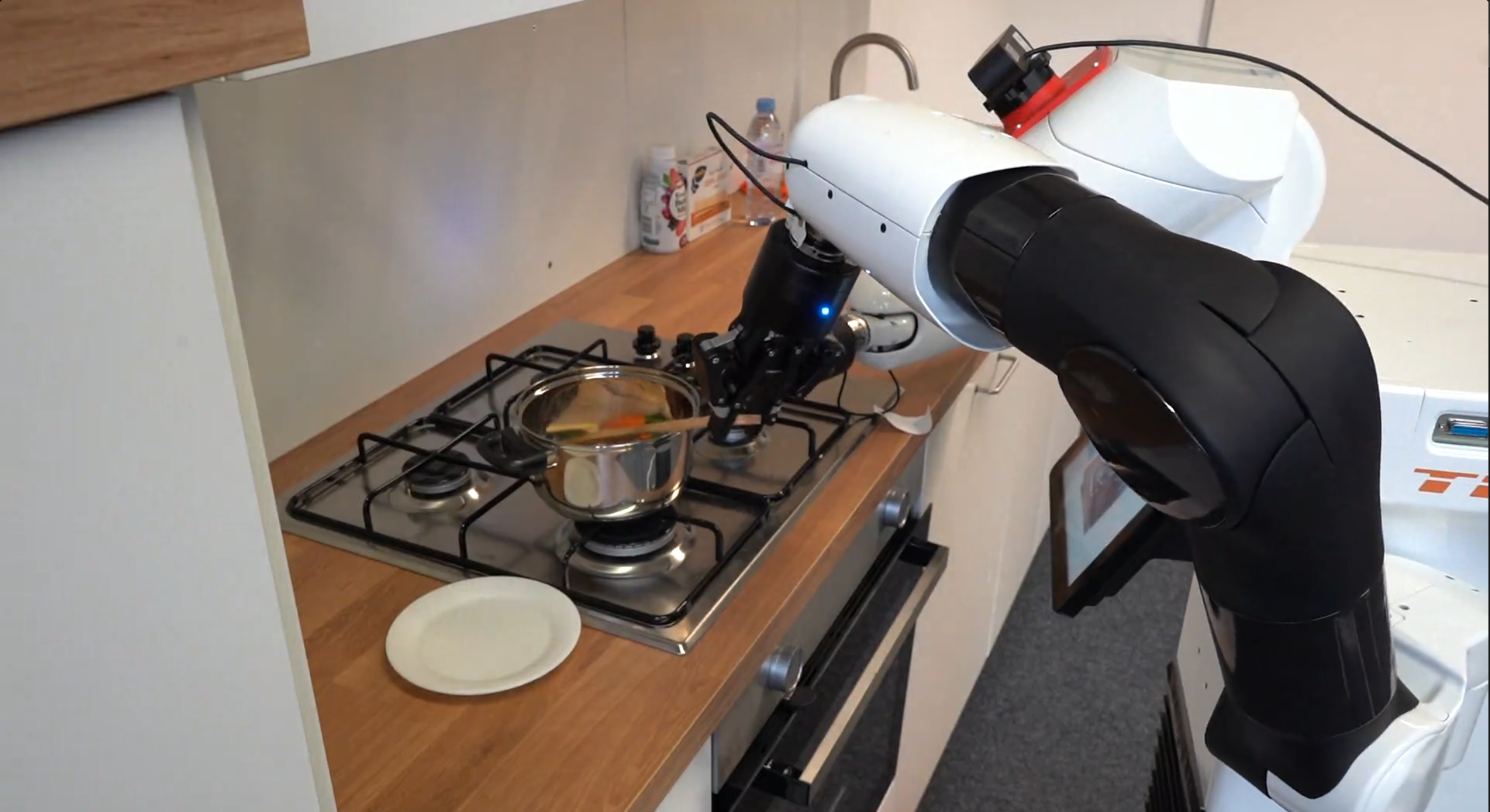}
    
    \caption{Demonstration how different affordance detections determine the success in performing a specific task.}
    \label{fig:robot_demo}
\end{figure}

Figure~\ref{fig:robot_demo} shows the robot performing the task of 'stirring vegetables'. The first example illustrates an unsuccessful attempt where the robot, relying on plain object segmentation, attempts to stir within the pot while holding the spatula too close to the middle. This suboptimal grip prevents the robot from reaching into the pot, making it impossible for it to complete the task of stirring the vegetables properly. The second example demonstrates an improvement, as the robot uses our affordance detection method to identify a better grasping region. However, it employs only one arm instead of two, leading to an unintended side effect since the pot is not stabilized. The stirring motion causes it to move, making the task more difficult.

In the final and most successful example, the robot fully utilizes both affordance regions detected by 2HandedAfforder. Here, the left end effector grasps the spatula closer to its edge while the right end effector holds the pot securely in place. This configuration enables a stable and effective stirring motion, demonstrating the advantages of incorporating our affordance predictions in bimanual robotic manipulation tasks.

\section{Qualitative Analysis of Mask Completion Approaches}
\label{sec:mask-completion-approaches}
\begin{figure*}[h!]
    \centering
    
    \begin{minipage}{0.3\textwidth}
        \centering
        \textbf{Original}  
    \end{minipage}%
    \hfill
    \begin{minipage}{0.3\textwidth}
        \centering
        \textbf{IR Based}  
    \end{minipage}%
    \hfill
    \begin{minipage}{0.3\textwidth}
        \centering
        \textbf{VS Based}  
    \end{minipage}

    \vspace{1em} 
    \centering
    \begin{minipage}{\textwidth}
        \centering
        \includegraphics[width=\textwidth]{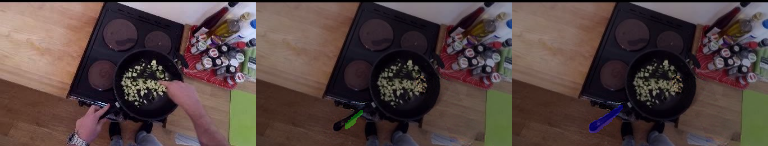} 
    \end{minipage}
    \begin{minipage}{\textwidth}
        \centering
        \includegraphics[width=\textwidth]{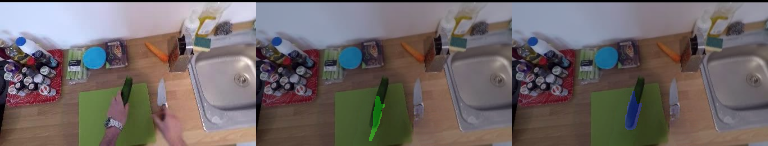} 
    \end{minipage}
    \begin{minipage}{\textwidth}
        \centering
        \includegraphics[width=\textwidth]{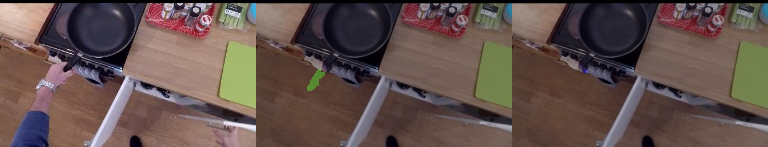} 
    \end{minipage}
    \begin{minipage}{\textwidth}
        \centering
        \includegraphics[width=\textwidth]{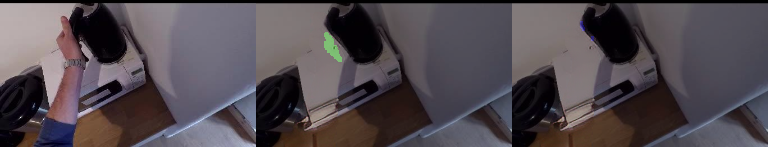} 
    \end{minipage}
    \begin{minipage}{\textwidth}
        \centering
        \includegraphics[width=\textwidth]{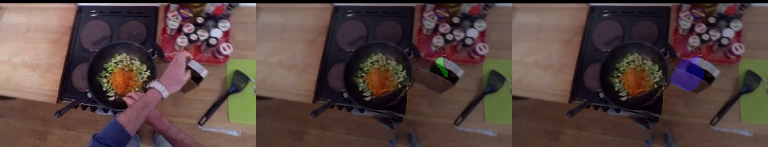} 
    \end{minipage}
    \caption{Examples of the two affordance extraction methods. The left column shows the original image, the center column shows the image reconstruction based approach and the right column shows the video segmentation based approach. The video segmentation based approach outperforms the image reconstruction based approach qualitatively in almost every instance.}
    \label{fig:2affmethods}
\end{figure*}

We developed and evaluated two mask completion approaches, i.e., an image reconstruction (IR) based and video segmentation (VS) based approach. The IR based approach uses the image inpainting model MI-GAN~\cite{sargsyan2023mi} to inpaint the missing regions of the mask using the hand mask as inpainting region. The VS based approach creates an image sequence out of the original image and the image with the hands removed. The object mask for the original image is then propagated to the inpainted image using SAM2~\cite{ravi2024sam} to create the completed version of the mask. The evaluation of these approaches was conducted qualitatively, and some examples can be seen in Fig.~\ref{fig:2affmethods}.
It is clearly visible that the VS based approach performs better on average than the IR based approach. Generally, the VS based approach provides more accurate results, see row 1 and 2, and it does not detect affordances at regions where the object was not reconstructed properly (row 3 and 4). This can be explained intuitively for two reasons:
The IR based approach has no information about the underlying RGB image and only processes the object mask itself which is binary by nature. So the image reconstruction model focuses on simple principles such as the continuation of lines and shapes. This leads to the reconstruction of what we call `ghost handles' where the image reconstruction model still predicts the existence of a handle or object part in general even though the object was not reconstructed successfully. This also reduces the accuracy of the IR based method. The VS based approach, however, has the information about the inpainted image and thus will only complete object parts that are properly inpainted. So, it naturally filters out data points where the object was not inpainted properly since the completed mask will not intersect the hand mask. Thus, it will always be more accurate and never predict `ghost handles'. There are only a couple of examples where the IR based approach performs better than the VS based approach (row 5).
Hence, we decided to use the VS based approach for the creation of 2HANDS.

\end{document}